\documentclass[a4paper,10pt]{article}

\usepackage{amsmath}
\usepackage{amssymb}
\usepackage{amsthm}
\usepackage{mathtools}
\usepackage{parskip}
\usepackage{graphicx}
\usepackage{psfrag}
\usepackage{subcaption}
\usepackage{url}

\usepackage{xtab,booktabs}

\usepackage{aligned-overset}
\usepackage{authblk}
\usepackage{color}
\usepackage{enumitem}  

\numberwithin{theo}{section}

\usepackage{xcolor}

\newcommand{\N}{\mathbb N}
\newcommand{\R}{\mathbb R}
\newcommand{\Sp}{\mathbb S}

\newcommand{\Z}{\mathbb Z}

\newcommand{\T}{\mathbb T}

\usepackage{bbm}

\usepackage{xtab,booktabs}

\usepackage{aligned-overset}
\usepackage{authblk}
\usepackage{color}
\usepackage{enumitem}  

\usepackage[hmargin=2.5cm,vmargin=2.5cm]{geometry}
\usepackage{appendix}

\title{Harnessing omnipresent oscillator networks as computational resource}

\author[a,1]{Thomas Geert de Jong}
\author[a]{Hirofumi Notsu}
\author[b]{Kohei Nakajima}

\affil[a]{Faculty of Mathematics and Physics, 
Institute of Science and Engineering,
Kanazawa University 
Kakuma, Kanazawa, 920-1192}
\affil[1]{Corresponding author: \texttt{tgdejong@se.kanazawa-u.ac.jp}}
\affil[b]{Graduate School of Information Science and Technology,
The University of Tokyo, Bunkyo-ku, Tokyo 113-0033}

\begin{document}

\maketitle




\begin{quote}
\noindent \textbf{Abstract:} Nature is pervaded with oscillatory dynamics. In networks of coupled oscillators patterns can arise when the system synchronizes to an external input. Hence, these networks provide processing and memory of input. We present a universal framework for harnessing oscillator networks as computational resource.  This computing framework is introduced by the ubiquitous model for phase-locking, the Kuramoto model. We force the Kuramoto model by a nonlinear target-system, then after substituting the target-system with a trained feedback-loop it emulates the target-system. Our results are two-fold.  
Firstly, the trained network inherits performance properties of the Kuramoto model, where all-to-all coupling is performed in linear time with respect to the number of nodes and parameters for synchronization are abundant. The latter implies that the network is generically successful since the system learns via sychronization. Secondly, the learning capabilities of the oscillator network,  which describe a type of collective intelligence, can be explained using Kuramoto model's order parameter. In summary, this work provides the foundation for utilizing nature's oscillator networks as a new class of information processing systems. \\

\noindent \textbf{Keywords:} collective intelligence, oscillator networks, synchronization, neuromorphic computing 
\end{quote}

\section{Introduction}
Oscillatory behaviour is foundational to the fabric of nature, ranging from pacemaker cells in the heart, to life-cycles of phytoplankton, to circadian rhythms, and to brainwaves~\cite{winfree1980geometry}. Typically, the dynamics arising from these phenomena is the result of a network of oscillators. When coupled, even weakly, and under the right conditions the oscillators adjust their phases such that the oscillator network exhibits synchronous behaviour~\cite{kuramoto1984chemical}. Such synchronous behaviour can be sustained without central guiding mechanism. Moreover, if the network is sufficiently complex the system is robust to damage, adaptable to change and scales to different sizes~\cite{lizier2023analytic}. Additionally, biological networks that we find in nature are efficient from an energy and resource perspective compared to their artificial counterparts~\cite{querlioz2024physics}. \\


We propose a framework for utilizing weakly coupled oscillatory networks as a computational resource to solve learning tasks such that the full system inherits properties of the coupled oscillators. 
 In this Article we apply the framework to the ubiquitous model for phase-locking, the Kuramoto model~\cite{kuramoto1975self}. More specifically, in the Kuramoto model the oscillators have different natural frequencies which for suitable coupling start rotating with a common frequency, i.e. the oscillators \textit{lock their phases}.  When adding external forcing the Kuramoto model can synchronize to external input~\cite{antonsen2008external,childs2008stability}. Hence, it provides processing and memory of the input.
We will consider a pattern prediction task where the pattern is a trajectory on a chaotic attractor which is described by a bounded deterministic yet unpredictable motion \cite{lorenz1963deterministic}. Chaotic attractors arise due to the geometry underlying the governing equations. Therefore, their complexities are quantifiable which gives additional performance tests for learning tasks. Moreover, like oscillatory phenomenon, chaotic dynamics are ubiquitous in nature~\cite{peitgen2004chaos}.

The computational capabilities of the Kuramoto model will be  harnessed using concepts of reservoir computing~\cite{jaeger2004harnessing, nakajima2020physical}. 
The coupled Kuramoto oscillators are subject to a forcing term driven by the input. Then decoupling the system from the input by substituting the input with a feedback-loop the network reproduces the attractor autonomously. Hence, this input-decoupled system is here referred to as the autonomous oscillator network but whenever it is clear from the context we shall refer to it as the oscillator network. \\

Our results are as follows:
\begin{itemize}
    \item[-] Although the system describes an all-to-all coupling, simulations inherit Kuramoto's linear time-complexity with respect to number of oscillators; 
    \item[-] The vast literature on Kuramoto models~\cite{strogatz2000kuramoto} can be used to explain the performance of the closed system. We use the so-called order parameter which averages the position of the oscillating components of the Kuramoto model and show that the time-series of the order parameter encodes the complex geometry of the attractor; 
    \item[-] The autonomous oscillator network is successful when a self-sustained synchronization is achieved. It is easy-to-configure in the sense that the hyper-parameters that lead to success are abundant. Additionally, for hyper-parameters that lead to failure we provide a simple test to identify oscillators that bar reproducing the attractor's geometry. 
\end{itemize}

\begin{figure*}[bt]
\centering
\includegraphics[width=15cm]{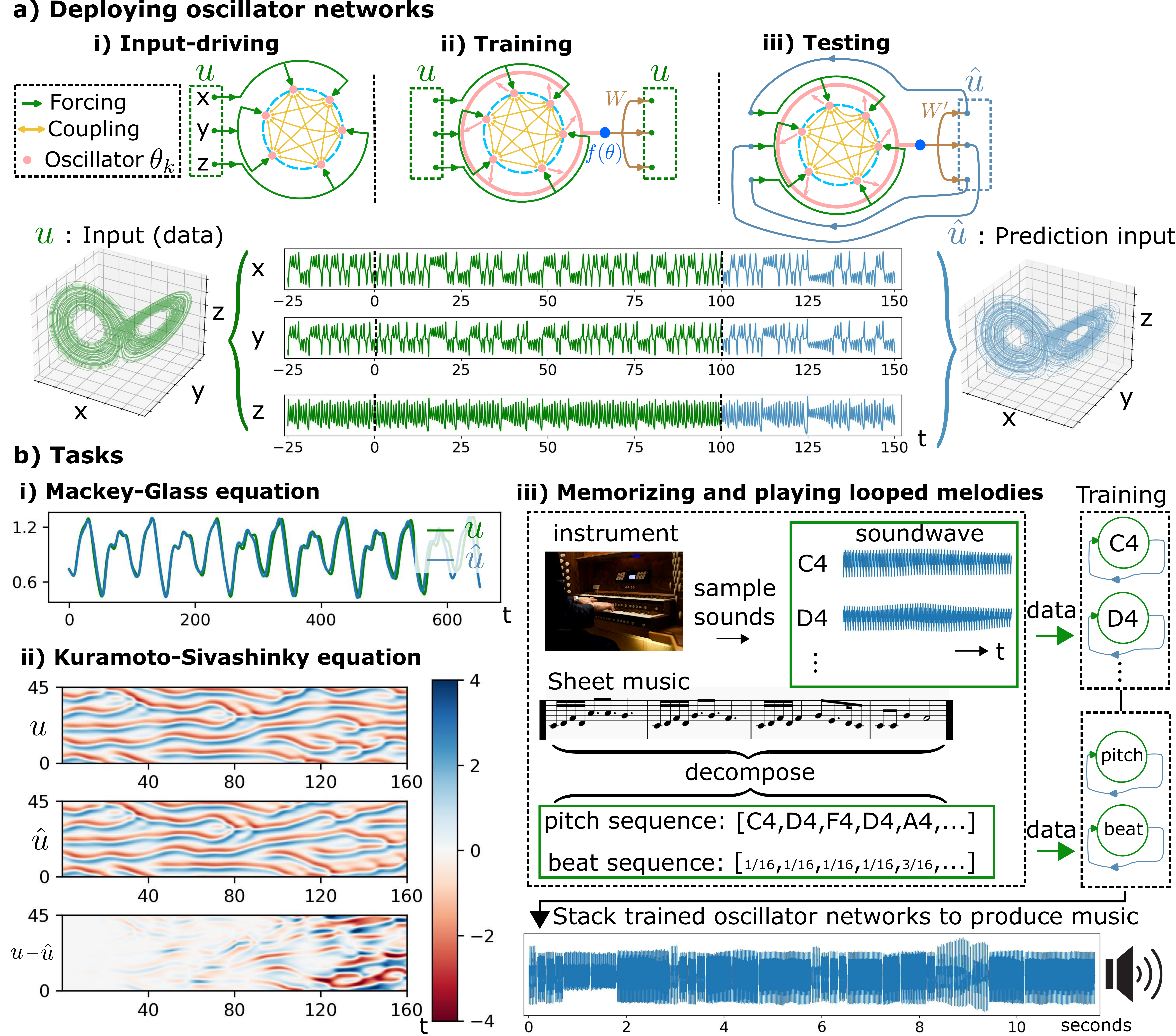}
\caption{\textbf{Overview for harnessing the Kuramoto model as computational resource:} (a) Deploying the oscillator network proceeds in three steps: (a-i) Input-driving, (a-ii) Training, (a-iii) Testing. (b) We present a variety of performance tasks for the autonomous oscillator network: (b-i) chaotic time delay differential system given by Mackey-Glass equation, (b-ii) infinite dimensional chaotic system given by Kuramoto-Sivashinky equation, (b-iii) memorizing and playing looped melodies by stacking oscillator networks. }
 \label{fig:overview_rc}
\end{figure*}

\section{Results}

\subsection{Background}

The modern study of collective synchronization was initiated by Winfree~\cite{winfree1967biological}. He cast the problem as a population of interacting limit-cycle oscillators. This problem proved to be complicated. Winfree realized that its complexity would be reduced if the couplings are weak and the oscillators nearly identical. Kuramoto~\cite{kuramoto1984chemical} continued this approach by realizing that using averaging techniques~\cite{sanders2007averaging} the long-term dynamics are given by the universal form:
\[
\frac{d \theta_k}{dt} =  \omega_k +  \sum_{j=1}^N \Gamma_{jk}(\theta_j-\theta_k), \;\; 1 \leq k \leq N,
\]
where $\theta_k(t) \in \R / 2 \pi \Z =: \Sp $ denotes the phase of the $i$th oscillator with $\omega_k \in \R$ its natural frequency and $\Gamma_{jk}$ the interaction function.  It will be useful to imagine that the oscillators $\theta= (\theta_1,\theta_2, \ldots, \theta_N)$ all move on a common circle. We will focus on the Kuramoto model for which $\Gamma_{jk}(\theta_j-\theta_k) = \frac{K}{N} \sin(\theta_j-\theta_k)$ with $K \geq 0$ the coupling strength.  Observe that this leads to an all-to-all coupling. Alternative $\Gamma_{jk}$ for our framework are discussed in the Supplementary Information.

\subsection{Kuramoto model as computational resource}

Harnessing the computational capabilities of the Kuramoto model proceeds in three steps (Fig.~\ref{fig:overview_rc}a).  The first step is to introduce a forcing term in the Kuramoto model that allows for synchronization to an external input $u$ (Fig.~\ref{fig:overview_rc}a-i). We consider an input ${u}(t)=(u_1(t), u_2(t), \ldots, u_M(t)) \in \R^M$.  The input-driven Kuramoto equations are then given by
\begin{align}
\frac{d \theta_k}{dt} =  \omega_k +  \underbrace{\frac{K}{N} \sum_{j=1}^N [\sin ( \theta_j - \theta_k ) ]}_{\rm coupling}  + \underbrace{F \sin ( c u_{v_k} - \theta_k)}_{\rm forcing},  \label{eq:gov} 
\end{align}
with $1 \leq  k \leq N$ and where $\omega_k$ will be sampled from  $2 \pi \mathcal{N}$ with $\mathcal{N}$ denoting the normal distribution with mean 1 and standard deviation 1, $v_k$ uniformly sampled from $\{1, 2, \ldots, M \}$, and $c$ the input scaling constant. The parameters $K,F \geq 0$ will be referred to as the coupling and forcing parameters, respectively. \eqref{eq:gov} is inspired by the Forced Kuramoto equations considered in~\cite{sakaguchi1988cooperative,antonsen2008external,ott2008low, childs2008stability}.

The second step is to train a function $g$ such that $g({\theta}) \approx {u}$ (Fig.~\ref{fig:overview_rc}a-ii). The aim is to use regression but as $\theta_k(t) \in  \Sp$ and $u_j(t) \in \R$ we need an intermediate function which maps  $\theta_k(t)$ to an appropriate topology. We define the $N$-dimensional torus $\T^N$ as the Cartesian product over $N$ sets of $\Sp$ and  consider the so-called read-out function $f: \T^N \rightarrow \R^{2N+1}$ given by $f({\theta}) = [1, \sin(\theta), \sin^2(\theta)]$ with the first component generating the bias-terms. Alternative read-out functions are considered in the Supplementary Information. Driving \eqref{eq:gov} for $T$ time-steps we represent the resulting time series vectors for $\theta$ and $u$ as an $N \times T$-matrix $\Theta$ and an $M \times T$-matrix $U$, respectively. We are then interested in determining
\begin{align*}
W' = \mathop{\arg \min}\limits_{W} \|  W \hat{f}(\Theta) - U \|_2^2  +  \varepsilon \| W \|_2^2,
\end{align*}
with $\hat{f}: \T^{N \times T} \rightarrow \R^{(2N+1)\times T}$  given by applying $f$ on the components in the time-direction and  where $\varepsilon$-term with $\varepsilon >0$ acts as a regularization term which will improve robustness. 

The final step (Fig.~\ref{fig:overview_rc}a-iii) is to substitute $u$ in \eqref{eq:gov} by $\hat{u}:= W'f(\theta)$,
\begin{gather}
\begin{aligned}
\frac{d \theta_k}{dt} &=  \omega_k + \frac{K}{N} \sum_{j=1}^N [\sin ( \theta_j - \theta_k ) ] + F \sin (c    \hat{u}_{v_k} - \theta_k   ).
\end{aligned} \label{eq:gov_cl}
\end{gather}
Then, we observe that $\hat{u} \approx u$. In this Article $\hat{u}$ will be referred to as the prediction. Since $u$ has been decoupled from \eqref{eq:gov_cl} this equation is referred to as the autonomous oscillator network. However, whenever it is clear from the context it will be referred to as oscillator network.  The autonomous oscillator network will be used for all the results in this Article.

\begin{figure*}[ht]
\centering
\includegraphics[width=14cm]{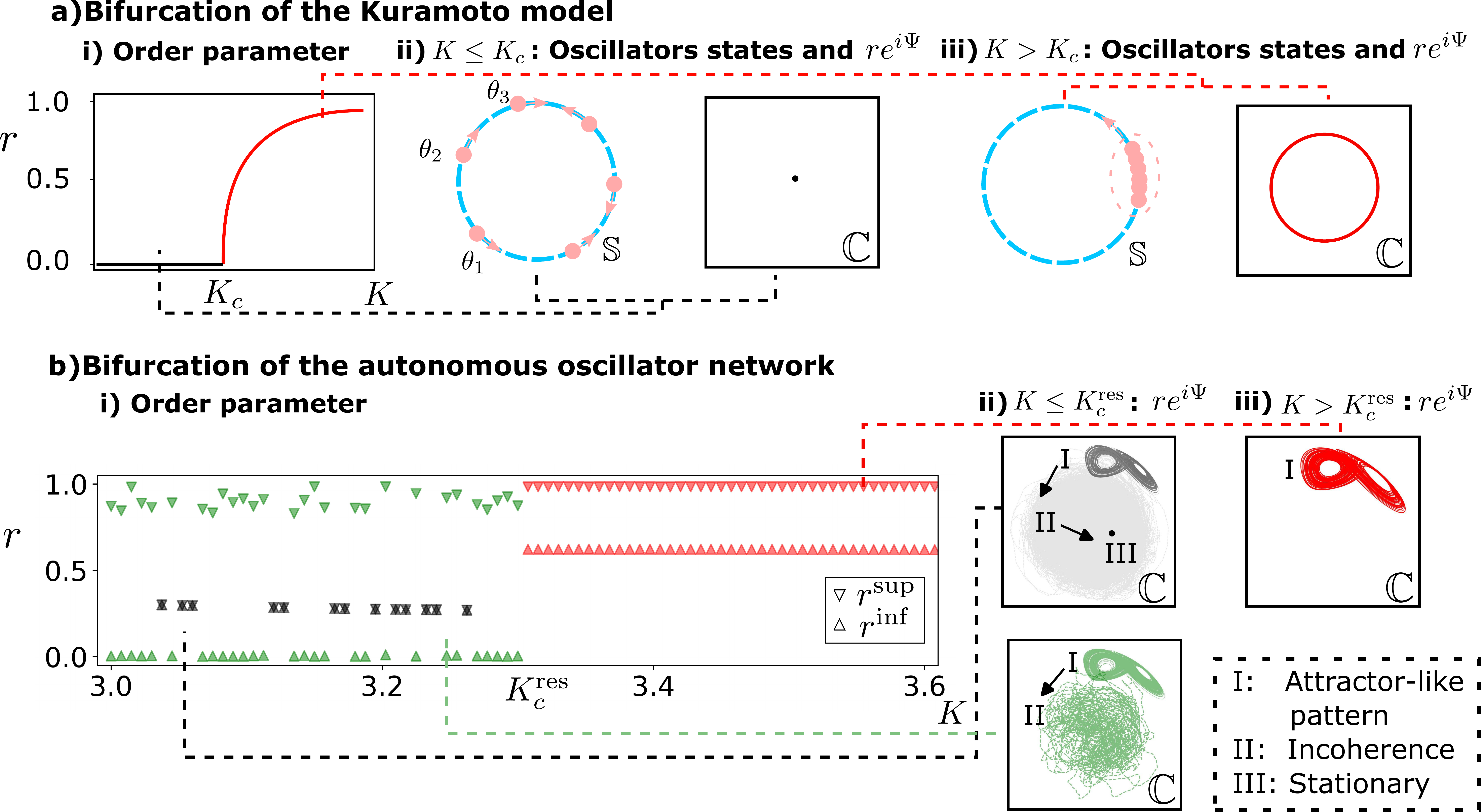}
\caption{\textbf{The oscillators undergo a bifurcation when the coupling is sufficiently large.} (a-i) In the Kuramoto model a bifurcation occurs in the order parameter $r$ when at a critical coupling $K_c$: (i) for $K \leq K_c$ no coherence occurs, (a-ii) for $K > K_c$ the system exhibits phase-locking. (b-i) In the autonomous oscillator network a bifurcation also occurs at a critical coupling $K^{\rm res}_c$: (b-ii) for $K \leq K^{\rm res}_c$ the dynamics in $re^{i\Psi}$ transitions from a Lorenz-like pattern to either a stationary state or indefinite incoherent dynamics,  (b-iii) for $K > K_c^{\rm res}$ the dynamics in $re^{i\Psi}$ indefinitely describes a Lorenz-like pattern.}  
\label{fig:kura_bifu}
\end{figure*}

Finally, we need to consider an input time series. We consider a trajectory $u = (x,y,z)$ on the Lorenz attractor which is obtained from
\begin{gather}
\begin{aligned}
\frac{dx}{dt} &= \sigma (y - x), \\
\frac{dy}{dt} &= x(\rho - z) - y,\\
\frac{dz}{dt} &= x y - \beta z ,
 \label{eq:lorenz}
\end{aligned}
\end{gather}
with the parameters $\sigma=10, \beta=8/3,\rho=28$ \cite{lorenz1963deterministic}. The stretching and folding around the wings of the Lorenz attractor gives rise to the chaotic dynamics which yields an unpredictable time evolution (Fig.~\ref{fig:overview_rc}a). We consider $N=1000$ all-to-all coupled oscillators and consider the input-driven system \eqref{eq:gov} for $-25 \leq t < 0$ (Fig~\ref{fig:overview_rc}a-i), then train the system for $0 \leq t < 100$ (Fig.~\ref{fig:overview_rc}a-ii), and finally switch to  \eqref{eq:gov_cl} for $t \geq 100$ (Fig.~\ref{fig:overview_rc}a-iii). In the Supplementary Information a detailed description is given of the numerical scheme. 

In (Fig.~\ref{fig:overview_rc}b) we consider additional benchmark tests. The parameter configurations can be found in the Supplementary Information. Furthermore, in (Fig.~\ref{fig:overview_rc}b-iii) we showcase memory capabilities of stacked autonomous oscillator systems by designing an architecture where independently trained networks are connected such that the collective can play looped melodies, see Movie S1,S2 for the oscillator evolution with audio. Finally, in the Supplementary Information we repeat all the experiments in this Article for another three-dimensional attractor given by the R\"{o}ssler system for the generality of our arguments. However, here we will continue with the Lorenz system.

\subsection{All-to-all coupling with linear complexity}

Observe that computation of the full vector field in \eqref{eq:gov} has $O(N^2)$-complexity. This is to be expected as the Kuramoto model describes an all-to-all coupling between oscillators. The so-called complex order parameter
proposed by Kuramoto~\cite{kuramoto1975self} is given by $\frac{1}{N} \sum_{k=1}^N e^{i \theta_k}$. Writing the complex order as
\begin{align}
r(t) {e}^{i \Psi(t)} = \frac{1}{N} \sum_{k=1}^N e^{i \theta_k(t)}, \label{eq:complex_order} 
\end{align}
with $r(t)\in \R,\Psi(t) \in \Sp$, the coupling term in \eqref{eq:gov} can be rewritten as
\begin{align}
\sum_{j=1}^N [\sin ( \theta_j - \theta_k ) ] = r \sin( \Psi - \theta_k). \label{eq:rpsi}
\end{align}
 Substitution of \eqref{eq:rpsi} into  \eqref{eq:gov_cl} makes the computation of the vector field $O(N)$-complexity.  The numerical method to solve \eqref{eq:gov_cl} preserves the linear complexity. 

 We note that in the simulations a slightly more efficient form of  \eqref{eq:complex_order} is considered, see the Supplementary Information.

\subsection{The bifurcation initiating learning}

The autonomous oscillator network is successful at the prediction task when the coupling parameter exceeds a critical value. The corresponding bifurcation can be understood via the complex order parameter \eqref{eq:complex_order}. But before continuing we review the classical bifurcation that occurs in the Kuramoto model, \eqref{eq:gov_cl} for $F=0$, using the complex order parameter.

The complex order parameter corresponds to the centroid of the phases and describes the collective rhythm produced by the oscillators~\cite{strogatz2000kuramoto}.
The radius $r$ determines the phase coherence of the oscillators and $\Psi$ is the average phase. For $r$ close to 0 the oscillators spread out over the circle and for $r$ close to 1 the oscillators are clumped together (Fig.~\ref{fig:kura_bifu}-ii,iii).

For $t \rightarrow \infty$ Kuramoto observed that  $r \rightarrow \rho_K$ for  $N \rightarrow \infty$  such that for $K \leq  K_c$ no coherence occurs as  $\rho_K=0$ (Fig. \ref{fig:kura_bifu}a-i,ii). But for $K > K_c$ we observe that $\rho_K >0$ as the oscillators exhibit phase-locking (Fig. \ref{fig:kura_bifu}a-i,iii). More specifically, for $K > K_c$ the order parameter rapidly increases to 1 as $K$ is increased .

In the autonomous oscillator network the bifurcation of $r$ with respect to $K$ differs from the bifurcation occurring in the Kuramoto model. In (Fig.~\ref{fig:kura_bifu}b) we fix $F$ for the oscillator network. We observe that there exists a critical parameter $K^{\rm res}_c$ such that for $K>K^{\rm res}_c$ the order parameter $r$ exhibits oscillating behaviour which in $r {e}^{i \Psi}$ appears Lorenz-like (Fig.~\ref{fig:kura_bifu}b-i,iii). For these parameters the prediction accurately approximates the target attractor for long-term evolution. For $K\leq K^{\rm res}_c$ the imprinting of the dynamics on the network becomes weaker and it transitions from Lorenz-like chaotic dynamics to a type of incoherence which corresponds to erratic motion in $r {e}^{i \Psi}$ (Fig.~\ref{fig:kura_bifu}b-i,ii). This incoherence can transition into the oscillators becoming stationary.

\subsection{Collective intelligence}

\begin{figure*}[ht]
\centering
\includegraphics[width=15cm]{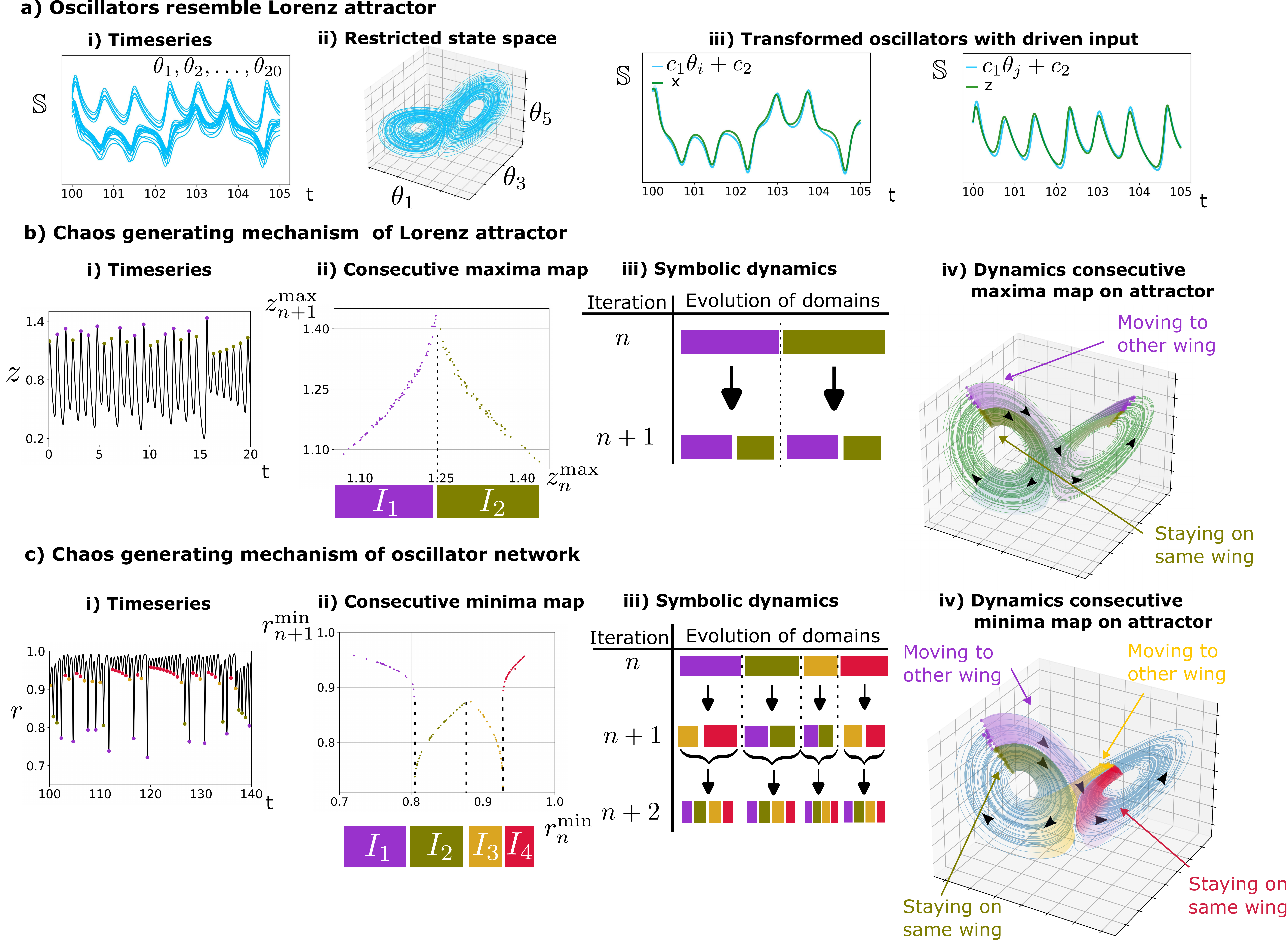}
\caption{\textbf{Chaos generating mechanism of the target attractor is collectively learned:} (a-i) The time-series of certain oscillators resemble the corresponding driven-input component which (a-ii) in a restricted state space can resemble the attractor. Under an affine transformation certain oscillators closely resemble the corresponding driven-input component. (b-i,ii) The chaotic dynamics in the Lorenz attractor can be exposed by investigating consecutive maxima, (b-iii,iv) the evolution of the domains describes the structure of the attractor. (c-i,ii) For the oscillator network the chaotic dynamics is exposed by investigating the map corresponding to consecutive minima of $r$, (c-iii,iv) the evolution of these domains also describes the structure of the attractor.}  \label{fig:symb}
\end{figure*}

Certain oscillators resemble the corresponding driven input component (Fig.~\ref{fig:symb}a-i). Furthermore, selecting three specific oscillators we obtain an attractor that looks Lorenz-like (Fig.~\ref{fig:symb}a-ii). Additionally, individual oscillators can closely resemble the Lorenz time series under an affine transformation (Fig.~\ref{fig:symb}a-iii). However, these results are rather unsatisfying as we are cherry-picking from $N=1000$ oscillators. For a more rigorous approach we require a method to study the oscillators as a whole. We will turn to $r$ in \eqref{eq:complex_order} as a tool for understanding the collective dynamics. 

Before continuing we present the classical procedure to expose chaos in the Lorenz system~\cite{lorenz1963deterministic}. Following the consecutive maxima of the $z$-variable (Fig.~\ref{fig:symb}b-i), $z_n^{\max}$, we observe that the map corresponding to $z_n^{\max} \mapsto z_{n+1}^{\max} $ corresponds to stretching and folding (Fig.~\ref{fig:symb}b-ii). The stretching implies that solutions start close to each other but separate after sufficiently long time, referred to as sensitive dependence on initial conditions. The folding ensures that the motion does not diverge. Separating the domains of the map into $I_1$ and $I_2$ we can symbolically represent the sensitive dependence (Fig.~\ref{fig:symb}b-iii) and then link the domains back to sections on the attractor (Fig.~\ref{fig:symb}b-iv).

We apply the previous procedure to the time series of $r$ for the autonomous oscillator network (Fig.~\ref{fig:symb}c-i). Instead of following consecutive local maxima of $r$ we consider consecutive minima, $r_n^{\min}$ (Fig.~\ref{fig:symb}c-ii). Studying the dynamics of the map corresponding to $r_n^{\min} \mapsto r_{n+1}^{\min} $. We can identify four domains, $I_1,I_2,I_3,I_4$ (Fig.~\ref{fig:symb}c-iii).  Applying the $r$ map twice these domains are being stretched and folded until they cover the full domain. Additionally, the domains can be uniquely identified to staying on a wing or moving to the other wing (Fig.~\ref{fig:symb}c-iv). This indicates that the oscillators collectively learn as they identify as a whole the underlying dynamics. We note that this procedure also works for local maxima, see the Supplementary Information. However, minima are chosen for aesthetic purposes as the maxima are clumped together.

\subsection{Forcing and coupling: the rise of synchronization and the collapse of internal geometry \label{sec:robust}}

By construction \eqref{eq:gov_cl} depends on the interaction between forcing and coupling. For the generalizability of the oscillator network it is paramount that the performance does not sensitively depend on the parameters. We will consider the short-term prediction and long-term dynamics to determine the configuration flexibility.

\begin{figure*}[ht]
\centering
\includegraphics[width=16cm]{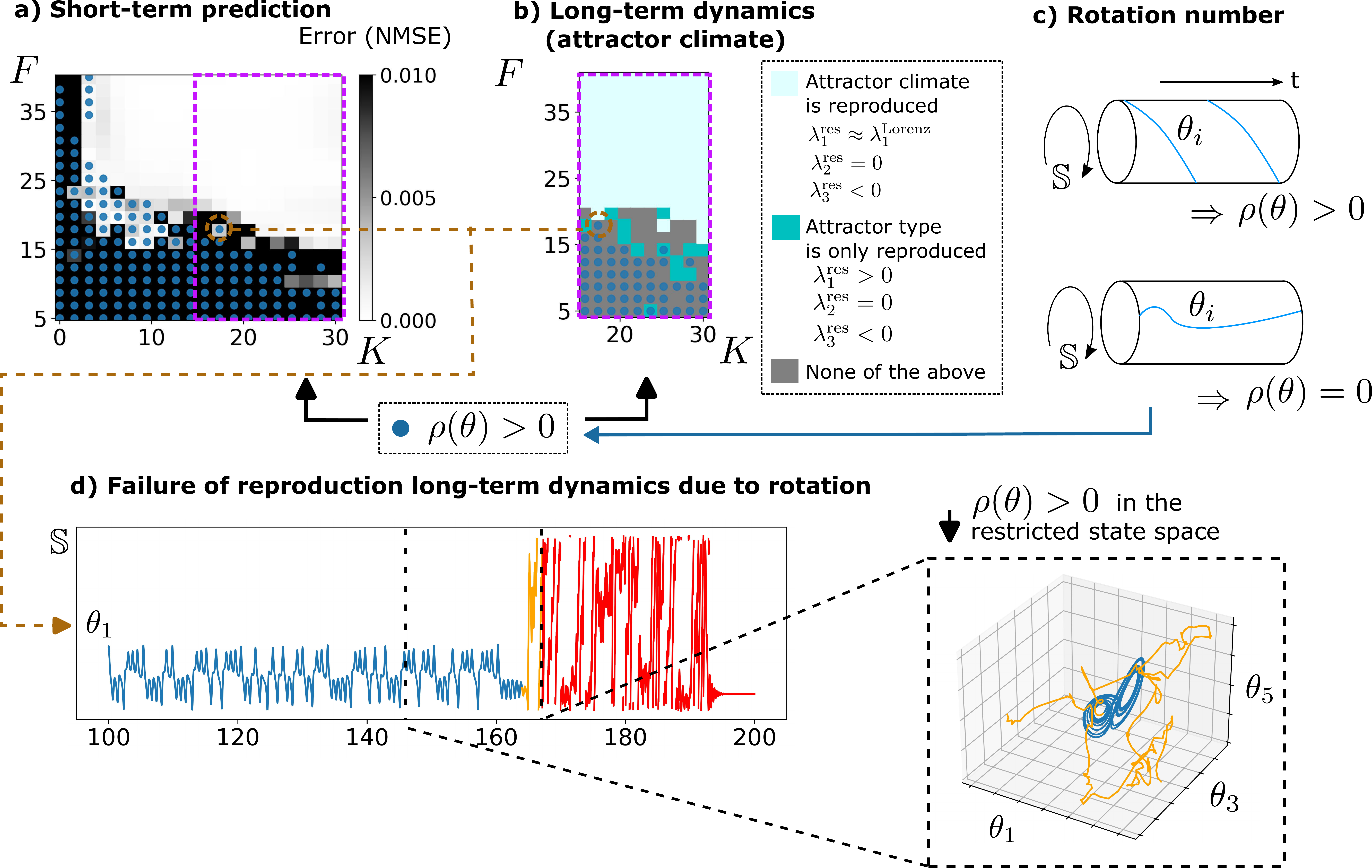}
\caption{\textbf{The connection between forcing, coupling, synchronization and internal geometry:} (a) Above a forcing and coupling threshold the short term prediction and (b) long term dynamics are accurate. (c) The geometry can be analyzed by computing the rotation number of the oscillators. Positive rotation number indicates the failure to reproduce the target attractor. (d) Rotations lead to destruction of the learned geometry from the target attractor. }
\label{fig:bifu} 
\end{figure*}

For the short-term prediction we use the Normalized Mean Square Error (NMSE) averaged over the components for the time duration of a couple of oscillations around the wings (the time interval [100,102]), see Supplementary Information. In (Fig.~\ref{fig:bifu}a) we observe that above a certain threshold of $F$ and $K$ we obtain high performance. Note that for $K=0$ the NMSE is large. This is to be expected since for $K=0$ the oscillators are not coupled while the Lorenz~\eqref{eq:lorenz} are coupled. Note that this shows that the learning capabilities of the collective are greater than the learning capabilities of the individual.

To study the long-term dynamics it is insufficient to consider an error measure on the prediction as the sensitive dependence on initial conditions of the Lorenz attractor ensures that solutions that start close to each other will separate. We use invariants that characterize chaos. Here, we compute Lyapunov exponents of \eqref{eq:gov_cl}~\cite{pathak2017using}. Lyapunov exponents measure the rate of separation for trajectories that start close to each other. The time interval considered for computing Lyapunov exponent depends on their convergence. However, typically we consider about 10 times the duration that was used in the generation of the Lorenz attractor (time interval [100,1100])  (Fig.~\ref{fig:overview_rc}). Denote the network's Lyapunov exponents by $\lambda_1^{\rm res} \geq \lambda_2^{\rm res} \geq \lambda_3^{\rm res} \geq \ldots$. The Lyapunov exponents for the Lorenz attractor are given by $\lambda^{\rm Lorenz}_1 \approx 0.91$, $\lambda^{\rm Lorenz}_2 =0$, $\lambda^{\rm Lorenz}_3 \approx -14.57$. For the long-term dynamics we restrict to chaotic attractors of the oscillator network which are of the same type as the Lorenz attractor, meaning that $\lambda_1^{\rm res} >0,\lambda_2^{\rm res} =0,\lambda_2^{\rm res}<0$. When  $\lambda_{1}^{\rm res} \approx \lambda_{1}^{\rm Lorenz}$ we say that the attractor's climate has been reproduced~\cite{pathak2017using}. We observe that the target attractor's climate is successfully reproduced when the forcing is sufficiently large (Fig.~\ref{fig:bifu}b). This fits our expectations since we would expect that synchronization improves with forcing. A more detailed Lyapunov exponent bifurcation diagram can be found in the Supplementary Information.

We note that the parameter $c$ in \eqref{eq:gov_cl}, the input/feedback-loop strength, was not considered in this section. But there is a parameter range for $c$ in which the oscillator network is successful. Additionally, from an application perspective we would expect that we can fully control the strength of the input/feedback-loop. When we consider sparse connectivity between the oscillators instead of all-to-all connectivity the successful region in $(c,F,K)$-parameter space greatly increases, see the Supplementary Information.

We conclude our results with investigating how the dynamics of \eqref{eq:gov_cl} can lead to situations where the geometry of the input is not learned or forgotten as the oscillators are evolved. Observe that the solutions of \eqref{eq:gov_cl} lie on the $N$-dimensional torus, $\T^N$, whereas the Lorenz attractor lies in $\R^3$. 
We define the rotation number for an oscillator $\theta_k$:
\begin{align}
\rho(\theta_k) =\left\lfloor \frac{1}{2\pi} \left| \int_{0}^{\infty} \theta_k'(\tau) d\tau  \right| \right\rfloor . 
\end{align}
Examples are given in (Fig.~\ref{fig:bifu}c). We check for which parameters there exists an oscillator satisfying $\rho(\theta_k)>0$. Returning to (Fig.~\ref{fig:bifu}b) we observe that $\rho(\theta_k)$ is a reasonable indicator for performance of short-term predictions with $\rho(\theta_k)>0$ indicating high error and $\rho(\theta_k)=0$ indicating low error. Then, returning to (Fig.~\ref{fig:bifu}c) we observe that climate reproducing oscillator networks satisfying $\rho(\theta_k)>0$ do not exist. 
Returning to (Fig.~\ref{fig:bifu}a) we observe that there are parameter values for which the indicator fails: $\rho(\theta_k)>0$ but the short-term prediction accuracy is high. In (Fig.~\ref{fig:bifu}d) we evolve these oscillators for multitudes of the short-term prediction duration. Initially the oscillator resembles the input time-series but after a while the oscillator starts rotating. During this rotation the prediction leaves the attractor-pattern.  As the oscillators are evolved the attractor-pattern is unraveled until it is forgotten and the oscillators become stationary.

\section{Discussion}

In this Article we presented a framework by which coupled oscillator networks can be harnessed as a linear time complexity, explainable, easy-to-configure, computational resource.  We have shown that the classical model for phase-locking, the Kuramoto model, can be harnessed to solve learning tasks. Since the Kuramoto model is a foundational model for a wide variety of applications this Article provides the evidence that the full range of Kuramoto-like systems have learning capabilities. Hence, this works offers a direct path for scientific communities that makes use of Kuramoto-like systems to utilize them in the context of artificial intelligence. 

Numerous works have shown the importance of synchronization in reservoir computers~\cite{lu2018attractor,verzelli2021learn,lymburn2019reservoir,weng2019synchronization}. This work puts forward a cornerstone by proposing that we can study synchronization capabilities of weakly coupled, nearly identical limit-cycle oscillators as a computational resource. Moreover, in the setting of Kuramoto oscillators we can use the technique from \cite{ott2008low} to formulate the continuum limit for $N \rightarrow \infty$ and study the synchronization capabilities analytically,  see the Supplementary Information.

There exist a variety of studies that have implemented Kuramoto-like systems for learning tasks. In~\cite{zuo2023self} pattern prediction tasks for a Kuramoto-like system is considered.  The dependency on the input is introduced in the distance between oscillators. Reminiscent of Hopfield networks the connectivity of the network is altered until it can reproduce the input. In ~\cite{chiba2024reservoircomputingkuramotomodel} a Kuramoto reservoir is presented with strong theoretical results, however experimentally it only predicts periodic inputs (sinus and triangle waves). In~\cite{shougat2022dynamic} a singular Hopf oscillator is considered which is used to generate network nodes artificially using a delay-operation called multiplexing. Although a multiplexed oscillator network could be investigated, it is preferable to increase the natural dimension of the oscillators as the information capabilities do not necessarily increase with respect to these artificial nodes.  Oscillators networks can be trained to perform classification tasks. This is achieved by training the oscillators to synchronize to specific inputs. In~\cite{vodenicarevic2017nanotechnology,vodenicarevic2018nano} this task is performed using a low dimensional Kuramoto-like model where the  natural frequencies are the weights and input of the network.

This framework could provide a foundational model for physical reservoir computers. The dynamics of Spin Torque Oscillators (STO) can be harnessed as a computational resource~\cite{torrejon17,markovic2019reservoir,furuta2018macromagnetic, tsunegi2019physical,tsunegi2023information}.  In~\cite{garg2021kuramoto,flovik2016describing} it is shown that the dynamics of coupled STOs can be modeled using a Kuramoto-like model. In \cite{baltussen2024chemical} the capabilities of a ground-breaking chemical reservoir computer is considered. In an idealized setting the chemical reservoir's governing equation can be modeled as a chemical reaction network. The rates are over a wide range of scales hence we conjecture that the oscillatory dynamics arises as the result of relaxation oscillations. These coupled oscillators are forced by a linearization of the forcing term in \eqref{eq:gov}. Consequently, this chemical reservoir computer could be seen as an implementation of our framework. \newpage

\appendix

\part*{Supporting Information}

\section{Introduction}

In Section \ref{sec:basics}-\ref{sec:lyabifu} we provide extensive supporting results (benchmarks, theoretical analyses, numerical schemes, explorations of future topics, etc.) to each result section in the main paper with the exception of the music task in (Fig. 1b-iii) of the main paper. We have devoted Section \ref{sec:music} to this task. In Section \ref{sec:rossler}, we reproduce all the experiments in the main paper for a system similar to the Lorenz system, the R\"{o}ssler system. 

\textbf{Note on the repository:} This work is accompanied by a repository which contains the code to run the experiments and generate the data as well as a database for parameter-configurations.  The repository is organized by the sections in the main paper and this document.

\section{Supplementary Information to Subsection 2.2 \label{sec:basics}}

\subsection{Deployment oscillator networks \label{sec:frame}}
We introduce the details of the framework and  notation that will be used throughout these materials. In Table \ref{tab:genparam} we present a summary of parameters. \\

We consider an input function $u: \R \rightarrow \R^M$. The governing equations take the form
\begin{align*}
\frac{d \theta_k}{dt} =  \omega_k +  
\sum_{j=1}^N \Gamma_{jk}(\theta_j-\theta_k)  + F \sin ( c u_{v_k} - \theta_k), \;\; 1 \leq  k \leq N  
\end{align*}
with $\theta_k \in \Sp := \R / 2 \pi  \Z $ and $\Gamma_{jk}: \Sp \rightarrow \R$ the interaction function. In the setting of Kuramoto oscillators we take $\Gamma_{jk}(\theta_j-\theta_k) = \frac{K}{N} \sin(\theta_j - \theta_k)$. The oscillator networks is deployed using the reservoir computing approach \cite{jaeger2004harnessing}: 
\begin{itemize}
\item[1.] \textbf{Wipe-out:} We consider the initial value problem corresponding to Equation \eqref{eq:gov} and compute the solution for all $t \in [-T_{\rm wipe},0)$.
\item[2.] \textbf{Training:}  We continue the solution for all $t \in [0,T_{\rm train})$.  We consider a yet-to-be-defined $f: \T^N \rightarrow \R^{N_{\rm ro}}$, referred to as read-out function. We find a  $W^{\rm out} \in \R^{M \times N_{\rm ro} }$ such that $W^{\rm out} f(\theta(t))$ predicts the input at the corresponding time. The numerical scheme will be discussed in \ref{sec:scheme}. Let us make the computation of $W^{\rm out}$ explicit. Assume $T_{\rm train}$ to be a multiple of $h_{\rm res}$ and define $n_{\rm train} := T_{\rm train}/h_{\rm res}$. We define 
$\Theta = (\theta(0), \theta(h_{\rm res}), \ldots, \theta(h_{\rm res}(n_{\rm train}-1)   ) \in \Sp^{N \times n_{\rm train}}$ and $U = (u(0), u(h_u), \ldots, u(h_u (n_{\rm train}-1)  ) ) \in \R^{M \times n_{\rm train}}$. Then we find 
\begin{align}
W' = \mathop{\arg \min}\limits_{W} \|  W \hat{f}(\Theta) - U \|_2^2  +  \varepsilon \| W \|_2^2, \label{eq:wprime}
\end{align}
with $\varepsilon>0$ the ridge-regression constant and  $\hat{f}: \T^{N \times n_{\rm train}} \rightarrow \R^{N_{\rm ro}\times n_{\rm train}}$  given by applying $f$ on the components in the time-direction. We compute $W'$ in \eqref{eq:wprime} using ridge regression~\cite{hoerl1970ridge}:
\[
W' =    U  \hat{f}(\Theta)^{\top} (\hat{f}(\Theta)\hat{f}(\Theta)^{\top}  + \varepsilon I)^{-1},
\]
with $I$ being an $N_{\rm ro} \times N_{\rm ro}$ identity matrix. 

\item[3.] \textbf{Testing:} In the governing ODE we substitute the input $u$ by $W^{\rm out} f(\theta)$. Let us denote the new dependent variable by $\hat{\theta}$. Then we compute the solution with initial value $\hat{\theta}(T_{\rm train}) ={\theta}(T_{\rm train})$ for all $t \in [T_{\rm train},T_{\rm train} + T_{\rm test})$ which is used to evaluate the performance of the oscillator network. More specifically, we compute $\hat{\theta}(t+(i-1) h_{\rm res})$,  evaluate $\hat{u }(t+(i-1) h_u):=W^{\rm out}f(\hat{\theta}(t+(i-1) h_{\rm res}))$ for $i=1,2, \ldots, n_{\rm test}$ with $n_{\rm test}= T_{\rm test}/h_{\rm res}$ and then compute the error between $\hat{u },u$ using the Normalized Mean Square Error (NMSE): 
\begin{align*}
{\rm NMSE} :=  \frac{1}{M} \sum_{j=1}^M \frac{\sum_{i=1}^{n_{\rm test}} | u_j(t_i)-\hat{u}_j(t_i) |^2}{ \sum_{i=1}^{n_{\rm test}} | {u}_j(t_i) |^2 } , \qquad t_i := T_{\rm train} + T_{\rm test}+ (i-1) h_u.
\end{align*}
Observe that the numerical scheme used to solve the $\hat{\theta}$-equation needs to take into account the discrete nature of the prediction map of $u$ given by $W^{\rm out} f(\hat{\theta})$. The scheme will be discussed in Section \ref{sec:scheme}.
\end{itemize}

\begingroup
 \begin{table}[ht]
\centering
\caption{Summary of parameters.}
 \label{tab:genparam}

 \begin{tabular}{ll}
\toprule
Symbol & Definition  \\
\midrule
$\Sp$ & $\R/2 \pi \Z $ \\
$\mathbb{T}^N$ &  $\underbrace{\Sp \times \cdots \times  \Sp}_{N-{\rm times}}$ \\
$F$ & forcing constant \\
$K$ & coupling constant \\
$c$ & input strength constant  \\
$\Gamma_{jk}$  & interaction function for $\theta_j$ and $\theta_k$ \\
$f$ & read-out function $f: \T^{N} \rightarrow \R^{N_{\rm ro}}$  \\
$N$ & number of oscillators \\
$M$ & input dimension  \\
$\varepsilon$ & ridge regression constant\\ 
$h_{\theta}$ & oscillator network time-step \\  
$h_{u}$ & input time-step  \\
 $T_{\rm wipe}$  & duration wipe-out phase given by multiple of $h_{\rm res}$ \\
 & (time-interval given by $[-T_{\rm wipe}, 0)$) \\
 $n_{\rm wipe}$ & $ \left\lfloor T_{\rm wipe} /h_{\rm res} \right\rfloor \in \N$ \\
$T_{\rm train}$ & duration training phase given by multiple of $h_{\theta}$ \\
& (time-interval given by $[0, T_{\rm train})$) \\
 $n_{\rm train}$ &  $ \left\lfloor T_{\rm train} /h_{\rm res} \right\rfloor \in \N$ \\
$T_{\rm test}$ &  duration test phase given by multiple of $h_{\theta}$\\
& (time interval given by $[T_{\rm train} , T_{\rm train} + T_{\rm test})$)  \\
 $n_{\rm test}$ & $\left\lfloor T_{\rm test} /h_{\rm res} \right\rfloor \in \N $\\
 \bottomrule
\end{tabular}
\end{table}
\endgroup

\subsection{Numerical scheme oscillator network \label{sec:scheme}}

The time-step $h_{\theta}$ is exclusively used by the states $\theta$ and the time-step $h_{u}$ is exclusively used by the input. Hence, for notational convenience we drop the subscripts of $h$.

We consider an input time series sampled with $\frac{1}{2h}$-frequency. Let $u^{(j)} = u(t_0+jh)$ with \\ $j=0,1/2,1,3/2,2, \ldots$.  We first consider the oscillator network during the wipe-out and training-stage. We will write the governing equations as $\frac{d \theta}{dt} = F(\theta,u)$ with $F$ denoting the vector field. The numerical approximation of $\theta(t_0+ih)$ is denoted by $\theta^{(i)}$ with $i=0,1,2 \ldots$. Then, we consider the Runge-Kutta 4th order (RK4) for the wipe-out and training phase:
\begin{gather}
\begin{aligned}
k_1 &= h F(\theta^{(i)}, u^{(i)}),\\
k_2 &= h F(\theta^{(i)}+k_1/2, u^{(i+1/2)}),\\
k_3 &= h F(\theta^{(i)}+k_2/2, u^{(i+1/2)}),\\
k_4 &= h F(\theta^{(i)}+k_3, u^{(i+1)}),\\
\theta^{(i+1)} & = \theta^{(i)} + (k_1+2k_2+2k_3+k_4)/6. 
\end{aligned} \label{eq:openrk4}
\end{gather}
During training we  compute a $g$ such that $g(\theta^{(i)}) \approx u^{(i)}$. We define $G(\theta^{(i)}) = F(\theta^{(i)}, g(\theta^{(i)}))$ and consider the following scheme for testing:
\begin{align*}
k_1 &= h G(\theta^{(i)}),\\
k_2 &= h G(\theta^{(i)}+k_1/2),\\
k_3 &= h G(\theta^{(i)}+k_2/2),\\
k_4 &= h G(\theta^{(i)}+k_3),\\
\theta^{(i+1)} &= \theta^{(i)} + (k_1+2k_2+2k_3+k_4)/6. 
\end{align*}

Another possible scheme for the testing phase is 
\begin{gather}
\begin{aligned}
k_1 &= h F(\theta^{(i)}, g(\theta^{(i)}))\\
k_2 &= h F(\theta^{(i)}+k_1/2, g(\theta^{(i)}))\\
k_3 &= h F(\theta^{(i)}+k_2/2, g(\theta^{(i)}))\\
k_4 &= h F(\theta^{(i)}+k_3, g(\theta^{(i)}))\\
\theta^{(i+1)} & = \theta^{(i)} + (k_1+2k_2+2k_3+k_4)/6 
\end{aligned} \label{eq:imprk4}
\end{gather}
This scheme can be interpreted as RK4 in $\theta$ and RK1 in $u$.

Observe that whenever $M \ll N$ these test phase schemes will run in $O(N)$.

\subsection{Oscillator network configuration for Lorenz system \label{sec:res_con}}

In Table \ref{tab:sum_not} we present the parameters used in the main experiment.

\begingroup
 \begin{table}[ht]
\centering
\caption{Summary of parameters for Lorenz prediction task. We note that $K,F$ are typically used as bifurcation parameters. Hence, the reported $K,F$ are only used in (Fig. 1) of the main paper. We note that the input has been scaled by 30.92 }\label{tab:sum_not}
 \begin{tabular}{ll}
\toprule
Symbol & Value \\
\midrule
$N$  & 1000 \\
$M$  & 3 \\ 
$F$ &  37.545 \\
$K$ & 20.680 \\
$c$ & 1.159 \\ 
$\varepsilon$  & $10^{-5}$\\
$h_{\theta}$   & $10^{-2}$ \\
$h_{\rm u}$   & $10^{-2}$ \\
 $T_{\rm wipe}$ & 25 \\
  $T_{\rm train}$   & 100 \\
 \bottomrule
\end{tabular}

\end{table}
\endgroup

The parameters were found using random search by sampling uniformly over parameter intervals.  The $c$ parameter appears to be important. Considering a value in $[0.9, 1.25]$ yields good results.  We rescaled the input variable by dividing the input by $30.92$. These values scale the input to a range in $[-\pi/2,\pi/2]$.  The $\omega_k$ are sampled from a normal distribution $2\pi \mathcal{N}(\mu,\sigma)$. The $\mu$ and $\sigma$ did not have a major impact on the performance so we just fixed them to 1. We fixed $N=1000$ as it is commonly used in the literature but we obtain similar performance for $N=500$. Additionally, $N=1000$ is commonly chosen in the Kuramoto literature for when the system exhibits continuum limit behaviour \cite{antonsen2008external}.

Let $\omega$ and $v$ denote the $N$-dimensional vectors corresponding to $\omega_k$ and $v_k$, respectively. We did not observe qualitatively different results for other $\omega$ and $v$.

\subsubsection{Data sets}
The Lorenz system was computed with RK4 using time-step $\Delta t = 1/2000$. We removed transients. We note that in all Lorenz system related figures in the main paper and this document the input has been divided by $30.92$ which is the divisor of $c$ divided by $10^{3}$.

\subsection{Read-out functions \label{supp:readout}}
We investigated the performance of the following read-out functions:
\begin{enumerate}
    \item $f(\theta) = [1, \sin(\theta) ] \in \R^{N+1}$,
    \item $f(\theta) = [1, \sin(\theta), \cos(\theta) ] \in \R^{2N+1}$,
    \item $f(\theta) = [1, \sin(\theta), \sin^2(\theta) ] \in \R^{2N+1}$.
\end{enumerate}
Performance was measured by sampling uniformly over all parameters, excluding $N,h,\mu,\sigma$ which were taken as in Section \ref{sec:res_con}, and then measuring NMSE of the prediction during testing. 

In Table \ref{tab:ro_perform} we arrange the performance from low to high. The read-out function 3. has most consistently high performance. We note that this read-out function is inspired by \cite{pathak2017using,lu2018attractor} in which linear and quadratic terms are considered. 

\begingroup
 \begin{table}[ht]
\centering
\caption{Performance of different read-out functions for $4 \cdot 10^3$ random samples over the parameter space.  }  \label{tab:ro_perform}
 \begin{tabular}{ll}
\toprule
Read-out & NMSE $<0.01$ \\
\midrule
$[1, \sin(\theta) ]$ &  0.05\% \\
$[1, \sin(\theta), \cos(\theta) ]$ & 30.10\% \\
$[1, \sin(\theta), \sin^2(\theta) ]$ & 52.45\% \\ 
 \bottomrule
\end{tabular}
\end{table}
\endgroup

\subsection{Comment on sampling natural frequencies from Cauchy distribution \label{sec:cauchy}}

Classically the Kuramoto model is considered with $\omega_k$ sampled from a Cauchy distribution:
\begin{align}
g(\omega) = \frac{\Delta_0}{\pi ( (\omega - \omega_{0})^2 + \Delta_0^2) },  \label{eq:cauchy}  
\end{align}
with $\omega_0, \Delta_0$ parameters. This is because the singularity of the distribution allows us to find an explicit expression in terms of an ODE for the order parameter when considering $N \rightarrow \infty$ \cite{ott2008low}. 

Cauchy distributions have flat tails compared to normal distributions. This appears to affect the performance of the oscillator network. More specifically, if we would consider $N=1000$, $\omega_0= \Delta_0=1$ then we are bound to find a couple of frequencies in $O(10)$. These large frequencies appear to negatively affect the performance. By considering  $\Delta_0$ much smaller such as $\Delta_0=0.01$ this problem can be circumvented.

\subsection{Additional benchmark tests \label{supp:bench}}
We consider additional benchmarks for the Kuramoto oscillator network. We used a primitive random search algorithm to discover suitable parameters. Hence, these parameters could still be improved on. The exact configurations can be found in the repository.

We consider input time-series generated by three different processes: Partial Differential Equation (PDE), time delay ordinary differential equation and Autoregressive Moving Average model (ARMA).

\textbf{a) Kuramoto-Sivashinsky (KS) equations:} We consider a PDE given by the fourth-order KS-equations  \cite{kuramoto1976persistent}:
$$ \frac{\partial y}{\partial t} + \frac{1}{2} \frac{\partial y^2}{\partial x} + \frac{\partial^2 y}{\partial y^2} + \frac{\partial^4 y}{\partial x^4} = 0, $$
on the periodic domain given by $y(t, 0) = y(t, L)$ with $L=45$. The input for the network is given by $u_i(t) = y(t, i \Delta x )$ with $\Delta x = L/M$, $i=1,2, \ldots,  M$. General oscillator network parameters are reported in Table \ref{tab:ks}. We note that here the performance depends strongly on $v_k$.

\begingroup
 \begin{table}[ht]
\centering
\caption{Summary of parameters for Kuramoto-Sivashinky: NMSE = 0.56}
 \label{tab:ks}
 \begin{tabular}{ll}
\toprule
Parameter & Value \\
\midrule
$N$ & 9000 \\
$M$ & 50 \\
$F$ & 47.27 \\
$K$ & 32.17 \\
$c$ &  0.271 \\
$h_{\rm res}$ & 0.01 \\
$h_{u}$ & 0.2 \\
$f(\theta)$ & $[1,\sin(\theta), \sin^2(\theta)]$ \\
$\varepsilon$  &  $10^{-5}$ \\
$n_{\rm wipe }$ & 5000  \\
$n_{\rm train}$ & $20 \cdot 10^3$ \\
$n_{\rm test}$ & 800 \\
 \bottomrule
\end{tabular}
\end{table}
\endgroup

\textbf{b) Mackey-Glass (MG) equations~\cite{mackey1977oscillation}:} The MG delay differential equation are given by :
$$
\frac{dy(t)}{dt} = \frac{a y(t - \tau)}{1+ y^n(t- \tau)} + b y(t),
$$
with the parameters $\tau=17$ and $(a,b,n)=(0.2,0.1,10)$. The input for the oscillator network is 1-dimensional and is given by $u(t) = y(t)$. General oscillator network parameters are reported in Table \ref{tab:ks}.  We note that to get the finer structure of the MG-equations we sampled $\omega_k$ from the normal distribution with $\mu=1,\sigma= 9$.

\begingroup
 \begin{table}[ht]
\centering
\caption{Summary of parameters for Mackey-Glass: NMSE = $9 \cdot 10^{-3}$}
 \label{tb:mg}

 \begin{tabular}{ll}
\toprule
Parameter & Value \\
\midrule
$N$ & 1000 \\
$F$ & 68.5 \\
$K$ & 52.2 \\
$c$ &  0.872 \\
$h_{\rm res}$ & 0.002 \\
$h_{u}$ & 0.816 \\
$f(\theta)$ & $[1,\sin(\theta), \sin^2(\theta)]$ \\
$\varepsilon$  &  $10^{-7}$ \\
$n_{\rm wipe }$ & 2200  \\
$n_{\rm train}$ & $10^{3}$ \\
$n_{\rm test}$ & 800 \\
 \bottomrule
\end{tabular}
\end{table}
\endgroup

\textbf{c) NARMA10~\cite{atiya2000new}:} Finally, we consider a nonlinear ARMA model. Given a uniformly random input $u^k\in[-1,1]$ the NARMA10 model is defined as
$$
y^{k+1} = \alpha y^k + \beta y^k\sum_{i=0}^{9}y^{k-i} + \gamma v^{k-9}v^k + \delta,
$$
$$
v^k=0.2 \frac{u^k+1}{2},
$$
where supscript $k$ indicates the $k$th time-step, $(\alpha,\beta,\gamma,\delta)=(0.3,0.05,1.5,0.1)$, and
the input $u^k$ is biased and scaled to $v^k \in [0.0, 0.2]$ to avoid divergence. Given $u^k$ and $y^k$ for $n_{\rm wipe} \leq k < n_{\rm train}$ the task is to predict $y^k$ given $u_k$ for $ n_{\rm train} \leq k <  n_{\rm test}$. General oscillator network parameters are reported in Table \ref{tab:narma}. An RK1-scheme is used because of the discontinuity of the input. We note that the test length used for the test error calculation is much longer than the test prediction given in the benchmark figure of the main paper. 

\begingroup
 \begin{table}[ht]
\centering
\caption{Summary of parameters for NARMA10: NMSE = $1.4 \cdot 10^{-3}$ }
\label{tab:narma}
 \begin{tabular}{ll}
\toprule
Parameter & Value \\
\midrule
$N$ & 500 \\
$F$ & 14.3 \\
$K$ & 1 \\
$c$ &  0.1 \\
$h_{\rm res}$ & 0.1 \\
$f(\theta)$ & $[1,\sin(\theta)]$ \\
$\varepsilon$  &  $10^{-11}$ \\
$n_{\rm train}$ & $2 \cdot 10^3$ \\
$n_{\rm test}$ & $2 \cdot 10^3$ \\
 \bottomrule
\end{tabular}
\end{table}
\endgroup

\subsubsection{Comparing performance to Echo State Network (ESN) for NARMA10}  

Let us consider the conventional (discrete) ESN \cite{jaeger2004harnessing,Jaeger2002,jaeger2001echo}. Let $x^k_i$ be the activation of the $i$th reservoir node at time-step $k$. Denote the reservoir weights connecting the $i$th node to the $j$th node by $w_{ij}$. The weights are uniformly randomly sampled from $[-1,1]$. However, a spectral radius $\rho$ is imposed on the resulting weight matrix. There are input weights connected to the $i$th input by $w^{\rm in}_i$ which are uniformly sampled from $[-\sigma, \sigma]$. The trainable weights will be denoted by $w^{\rm out}_i$ which will be obtained using ridge regression. The discrete time evolution is given by
\begin{align*}
x^k_i &= f\left(\sum_{j=1}^N w_{ij}x_{j}^{k-1}+ w^{\rm in}_i u^k \right) , \\
\hat{y}^k &= \sum_{i=0} w^{\rm out}_i x_{i}^k,
\end{align*}
where we take $x^{k}_0=1$ so that the corresponding weight becomes a bias term and we consider $f: \R \rightarrow \R$ given by $f(x) = \tanh(x)$.  

Observe that evolving ESN has $O(N^2)$-complexity and recall that computation of the vector field for the Kuramoto oscillator network has $O(N)$-complexity. Hence, if we want to compare the Kuramoto oscillator network with configuration given by Table \ref{tab:narma} we need to consider an RK1-scheme for the Kuramoto oscillator network and ESN with $N=23$. For the NARMA task we perform a grid search for ESN using the methodology described in Appendix B of \cite{nakajima2019boosting}.  We find that ESN achieves NMSE = $7.9 \cdot 10^{-4}$ which is significantly smaller than the Kuramoto oscillator network which achieves NMSE $1.4 \cdot 10^{-3}$. However, for ESN with $N=10$ we obtain NMSE = $1.6 \cdot 10^{-3}$ and for the Kuramoto oscillator network with $N=100$, configuration as in Table \ref{tab:narma} and with 2-decimal optimized $c$, which is $c=0.15$, also gives NMSE = $1.6 \cdot 10^{-3}$. 

Let us note that this is not a fair comparison because NARMA10 is a discrete task. Hence, a continuous system such as the Kuramoto oscillator network is expected to perform worse than a discrete reservoir such as conventional ESN. Additionally, the coupling strength in the Kuramoto oscillator network is taken constant whereas ESN has random internal weights. Hence, from a connectivity perspective ESN is richer than the network of the Kuramoto oscillator network. Of course richer network connectivity classes can be considered for more general oscillator networks, see Section \ref{sec:kura_var}.

\subsubsection{Comments on initial condition} 

Typically the initial condition is chosen as $\theta_i(-T_{\rm wipe}) = 2\pi (i-1)/(N-1)$. For $N$ large this means in terms of the complex order parameter, $z = r e^{i \Psi}$, that $0<|z(-T_{\rm wipe})| \ll 1$. The complex order parameter for initial conditions sampled uniformly from $[0,2\pi)$ will have similar properties and similar performance. Additionally, it appears that when the embedded attractor of the oscillator network in states space is represented in the complex order parameter it is located in the right half-plane. Observe that $z(-T_{\rm wipe})$ is also located in the right half-plane. 

We observe that there is a significant dip in performance for MG and NARMA10 when we consider randomly sampled initial conditions. However, this can be resolved by considering a longer wipe-out phase, $n_{\rm wipe} = 5000$ which is significantly larger than what is used for ESN in the literature. Averaging over 10 uniformly sampled random initial conditions we obtain for MG that NMSE = $\mu \pm \sigma = 1.1  \cdot 10^{-2} \pm 7 \cdot 10^{-3}$ and for NARMA10 that NMSE = $\mu \pm \sigma  = 2.0 \cdot 10^{-3} \pm 8 \cdot 10^{-4} $. For KS we already considered $n_{\rm wipe} = 5000$ and the NMSE barely varies for random initial conditions.

\subsubsection{General comments on improving performance}

The Kuramoto oscillator network as introduced in the main paper is the \textit{vanilla} oscillator network. There are a number of aspects that can be configured to improve performance: distribution for natural frequencies (for example, in the case of MG a bimodal distribution appears to improve performance, also see Section \ref{sec:cauchy}), interaction function, oscillator connectivity. The last two points will be discussed in the next section.

\subsection{Kuramoto oscillator network variations \label{sec:kura_var}}

In this section we consider Kuramoto-like systems to which we apply the framework as presented in \ref{sec:frame}. More specifically, we consider Equation \eqref{eq:gov} for different $\Gamma_{jk}$. The first is the Kuramoto-Sakaguchi model \cite{sakaguchi1986soluble}: 
\begin{align}
\Gamma_{jk}(\theta_j - \theta_k) = \frac{K}{N} \sin(\theta_j - \theta_k + \alpha). \label{res:saka}
\end{align}
We consider the case $\alpha=0.1$ and obtain similar performance result compared to the Kuramoto oscillator network we do not report on the details and refer to the accompanied code. Kuramoto-Sakaguchi model is a more general form of the Kuramoto model. Hence, this supports generalizability of the framework.

The Kuramoto model describes all-to-all coupling. Instead we can consider connectivity described by an arbitrary graph: 
\begin{align}
\Gamma_{jk}(\theta_j - \theta_k) = a_{jk} \sin(\theta_j - \theta_k )
\label{res:graph}
\end{align}
with $A=(a_{jk}) \in \R^{N \times N}$ the adjacency matrix of the graph describing the connectivity between $\theta$. Let us explore the performance based on the connectivity of $A$. We sample $A$ from a graph model, the parameters $c,F,K$ from a uniform distribution, $\omega \in 2\pi \mathcal{N}$ and then investigate the corresponding performance distributions in NMSE. All remaining parameters are fixed and given by the values in Table \ref{tab:sum_not}. For the data set we consider the Lorenz time-series with $n_{\rm test} = 200$. We will sample $A$ from the following graph models:
\begin{itemize} 
    \item[-] $N$-node Erd\"{os}-Renyi random graph model \cite{bollobas1998random} with each edge included in the graph with probability $p=0.1$;
    \item[-] $N$-node regular graph model with degree $100$;
    \item[-] $N$-node Watts-Strogatz model \cite{watts1998collective}, random graph model with small-world properties, with mean degree $m=100$ and edge-rewiring probability $p=0.5$.   
\end{itemize}
We note that the regular graph model is not a random graph model. Note that the vertex degree distribution of the random graph model has mean $99.9$. which is close to the mean degree of the regular graph model and Watts-Strogatz model.  The performance results are displayed in Table \ref{tab:graph}. The random graph models have the best performance. Notably the regular graph model already outperforms the vanilla Kuramoto oscillator network given by the all-to-all connected graph.   We also performed experiments for connected graphs sampled from Watts-Strogatz. The performance for these connected graphs yields similar performance to Erd\"{o}s-Renyi over the ranges considered in Table \ref{tab:graph}. 

\begingroup
 \begin{table}[ht]
\centering
\caption{Performance of short-time prediction for different graph models: results precise up to $\pm 1 \%$ } \label{tab:graph}
 \begin{tabular}{lcccc}
\toprule
 & all-to-all  & Erd\"{os}-Renyi & regular  & Watts-Strogatz  \\
\midrule
NMSE $<10^{-2}$ & 43\% & \textbf{73\%} & 68\%  & \textbf{73\% } \\
NMSE $<10^{-3}$ & 22\% & \textbf{63\%} & 52\%  & 58\% \\
NMSE $<10^{-4}$ & 4\% &  \textbf{27\%} &  20\% &  \textbf{27\%}\\ 
 \bottomrule
\end{tabular}
\end{table} 
\endgroup

\section{Supplementary Information to Subsection 2.3 \label{sec:lino}}

\subsection{Hardware and computational speed}

The CPU used for the experiments is an AMD Ryzen 9 5950X (16-core). The code was written in Python and the module Numba was used as an acceleration tool. Wipe-out, training and testing  for a test length of $2 \cdot 10^5$ with the parameters in Table \ref{tab:sum_not} can be performed in approximately 15 seconds. The procedures in \cite{de2023mlncp} are used to accelerate the computations.

\subsection{$O(N)$ vector field used in experiments \label{supp:on}}

In none of the experiments we need to explicitly compute $\Psi$. So it makes more sense to write
\begin{align}
\sum_{j=1}^N \sin ( \theta_j - \theta_k )  =  R_1(\theta) \cos(\theta_k) - R_2(\theta) \sin( \theta_k),
\end{align}
where $R_1(\theta) =  \sum_{j=1}^N \sin( \theta_j)$, $R_2(\theta) =  \sum_{j=1}^N \cos( \theta_j)$. Substitution into \eqref{eq:gov} makes the computation of the vector field $O(N)$-complexity.

\section{Supplementary Information to Subsection 2.4 \label{sec:learn}}

\subsection{Bifurcation in $K$ for fixed $F$: increasing $\varepsilon$}

\begin{figure}[ht]
    \centering
    \includegraphics[width=15cm]{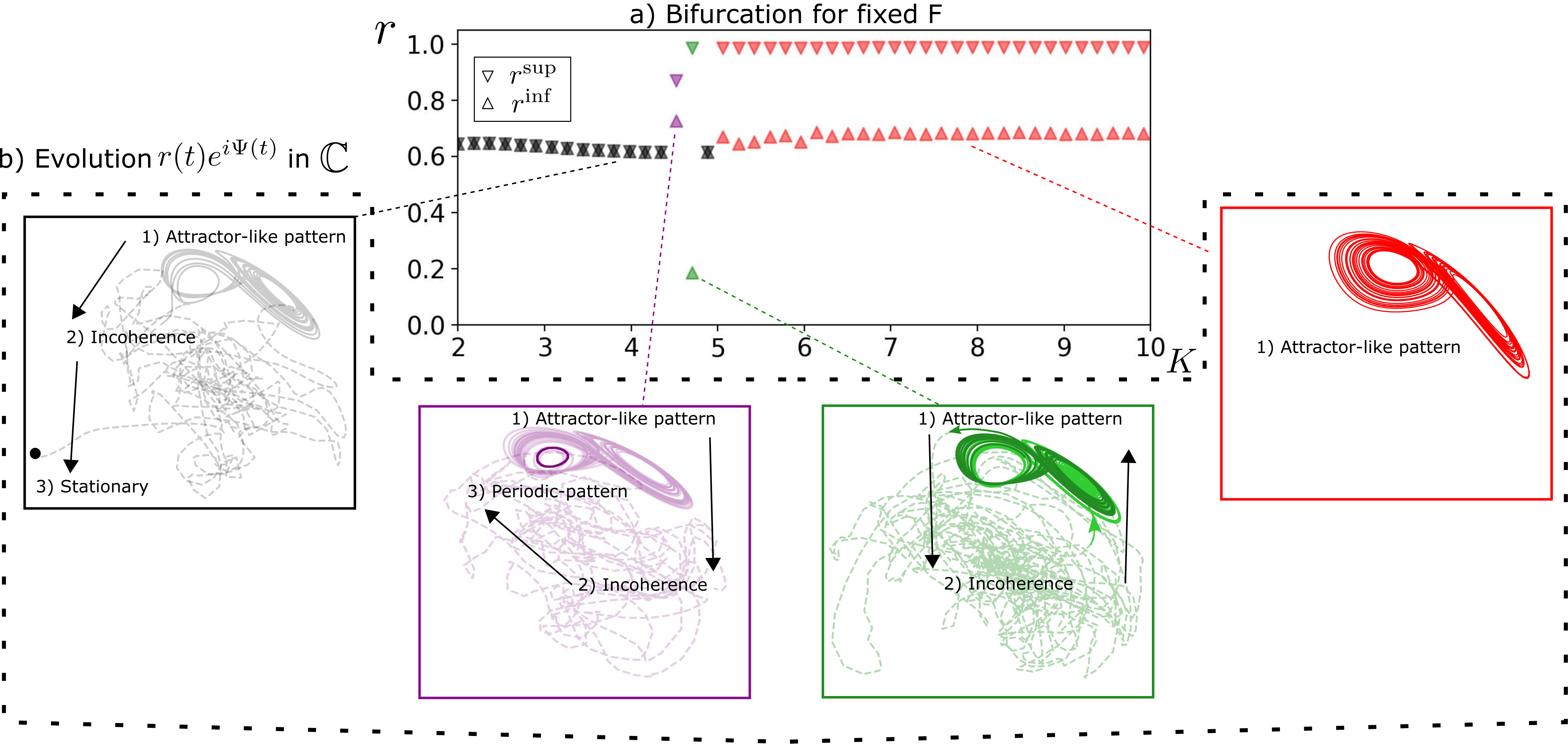}
    \caption{Bifurcation in $K$ for $F=35$ with $\varepsilon=10^{-3}$}
    \label{fig:bifuK_reg}
\end{figure}

When we increase the ridge regression constant from $\varepsilon=10^{-5}$, used in the main paper (Table \ref{tab:sum_not}), to  $\varepsilon=10^{-3}$ new dynamics appear in the learning bifurcation diagram (Fig. \ref{fig:bifuK_reg}). Recall that we defined the critical parameter as the smallest $K$ for which the order parameter exhibits long-term Lorenz-like dynamics. We observe that the critical parameter for $\varepsilon=10^{-5}$ is around 3.3 which is smaller than than the critical parameter for $\varepsilon=10^{-3}$ which is around 4.8 (Fig. \ref{fig:bifuK_reg}a). 

Comparing (Fig. \ref{fig:bifuK_reg}a) to the bifurcation in the main paper we observe that a transition to a permanent incoherent state does not occur. However, in (Fig. \ref{fig:bifuK_reg}b) we do observe that the motion can alternate between attractor-like pattern and incoherence. Notably decreasing $K$ we obtain a configuration where the complex order parameter converges to periodic motion inside the lobes of the Lorenz-like wings, the purple box (Fig. \ref{fig:bifuK_reg})

\subsection{Bifurcation in $F$ for fixed $K$ \label{sec:bifuF}}
Returning to the parameters considered in the main paper (Table \ref{tab:sum_not}) we now consider the bifurcation in $F$ for fixed $K$.

\begin{figure}[hbt!]
    \centering
    \includegraphics[width=15cm]{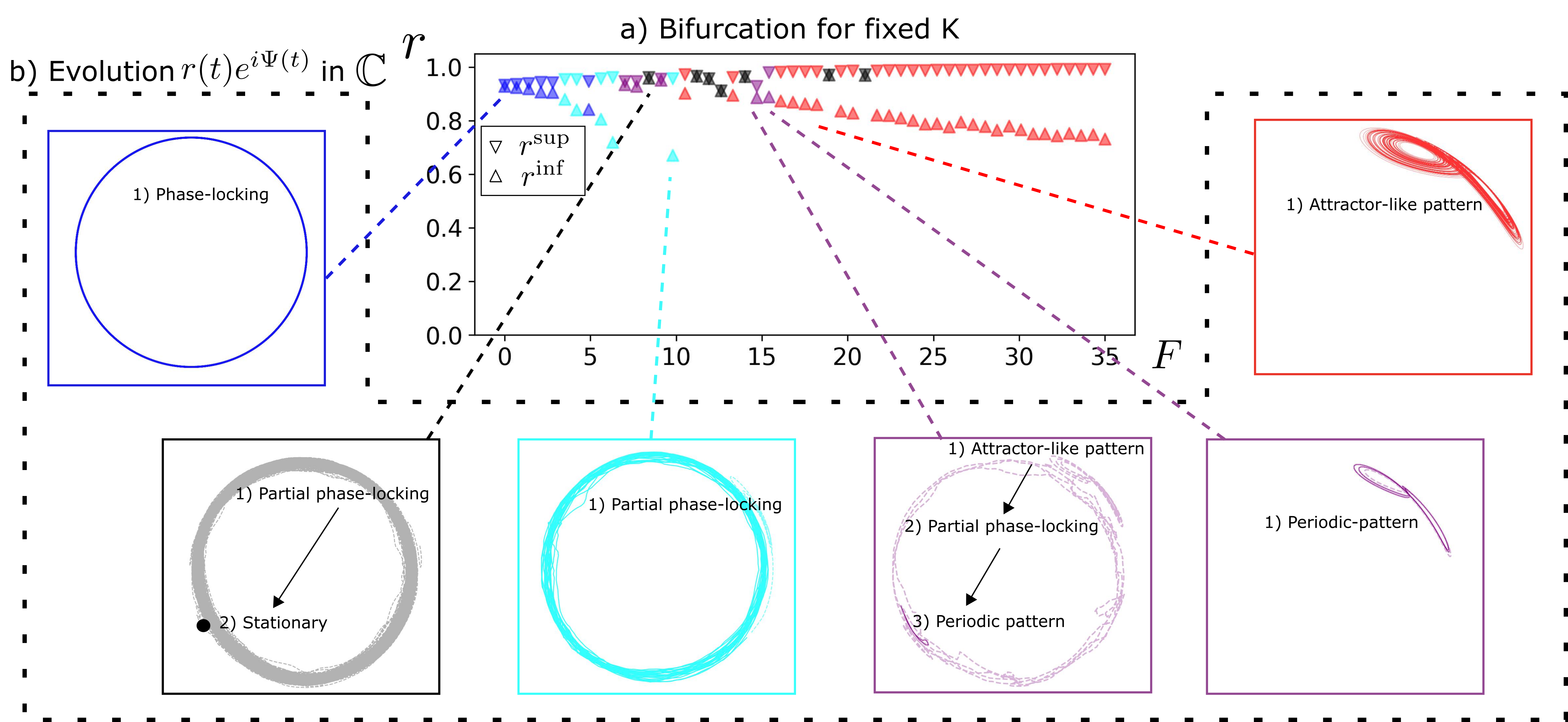}
    \caption{Bifurcation in $F$ for $K=20$ with $\varepsilon=10^{-5}$}
    \label{fig:Fbifu}
\end{figure}

 In short the bifurcation occurs as the result of the phase-locking state losing its stability for sufficiently large $F$. Although periodic motion of the complex order parameter can occur for a range of $F$ (Fig. \ref{fig:Fbifu}) when $F$ is sufficiently large the complex order parameter converges to a Lorenz-like attractor.

\subsection{Numerical issues related to sensitive dependence on initial conditions}

The bifurcations in this section require the study of the long-term dynamics of solutions.
To accelerate the computational speed throughout the paper the Numba module is used \cite{lam2015numba}. When the code is compiled, Numba can change the order of computations to optimize the computational speed. Hence, depending on the Numba version the quantitative properties of the solutions can be affected. This specifically concerns the transition times of the solutions in the bifurcation diagrams of this section. 

\subsection{Continuum limit Kuramoto oscillator network}

\subsubsection{Classical forced Kuramoto equations}
The governing equation draw inspiration from the classical forced Kuramoto equations:
\begin{align}
\frac{d \theta_k}{dt} =  \omega_k +  \frac{K}{N} \sum_{j=1}^N \sin ( \theta_j - \theta_k )   +  F \sin ( ct - \theta_k) .\label{eq:forcedkura}
\end{align}
In the context of the main article we have that $u: \R \rightarrow \T $ given by $u(t) = ct$.
\begin{figure}[hbt!]
\centering
	\begin{subfigure}[b]{0.45\textwidth}
	\centering
	\includegraphics[width = 6cm]{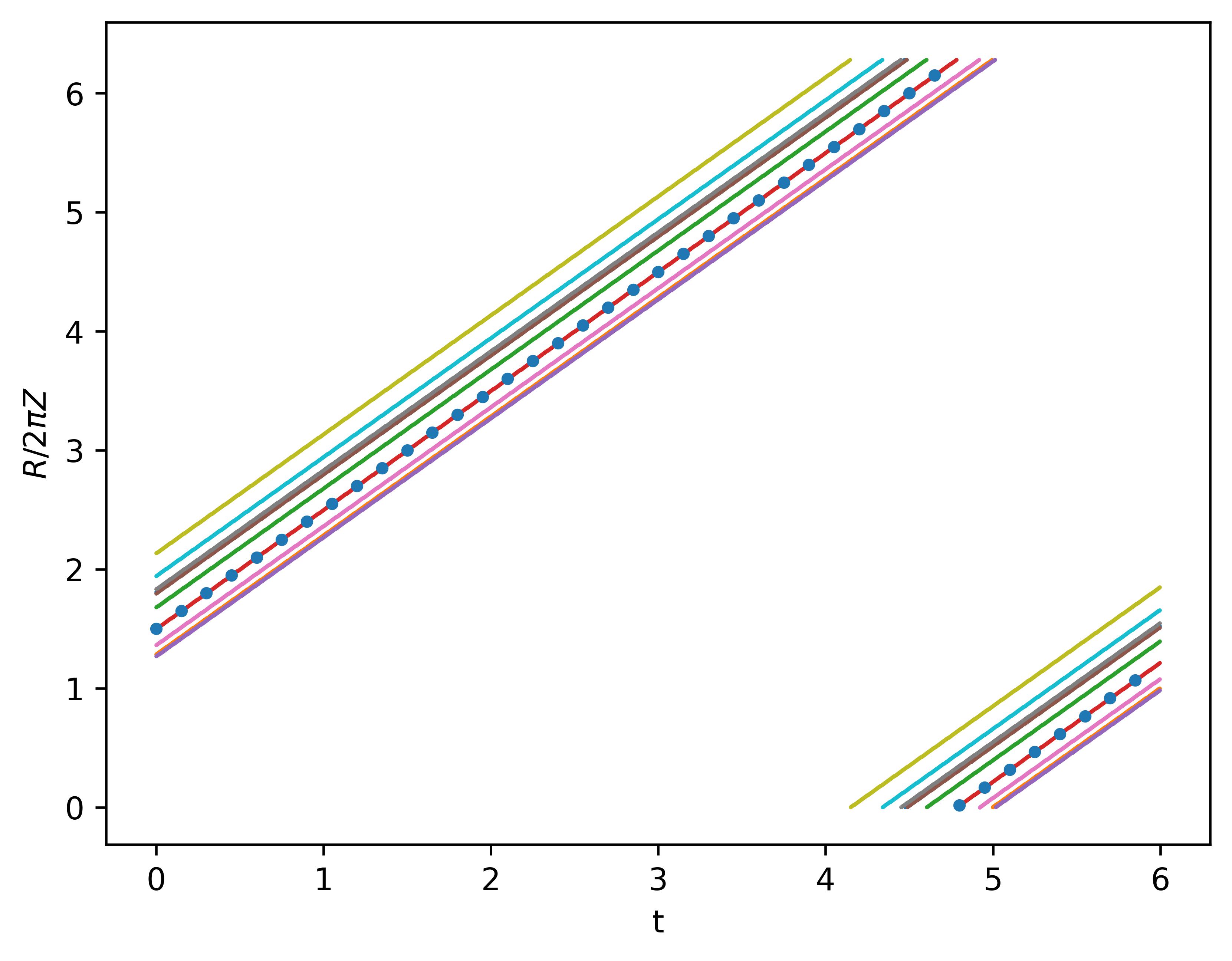}
        \caption{$F=15$}
	\end{subfigure}
    \begin{subfigure}[b]{0.45\textwidth}
	\centering    
    \includegraphics[width= 6cm]{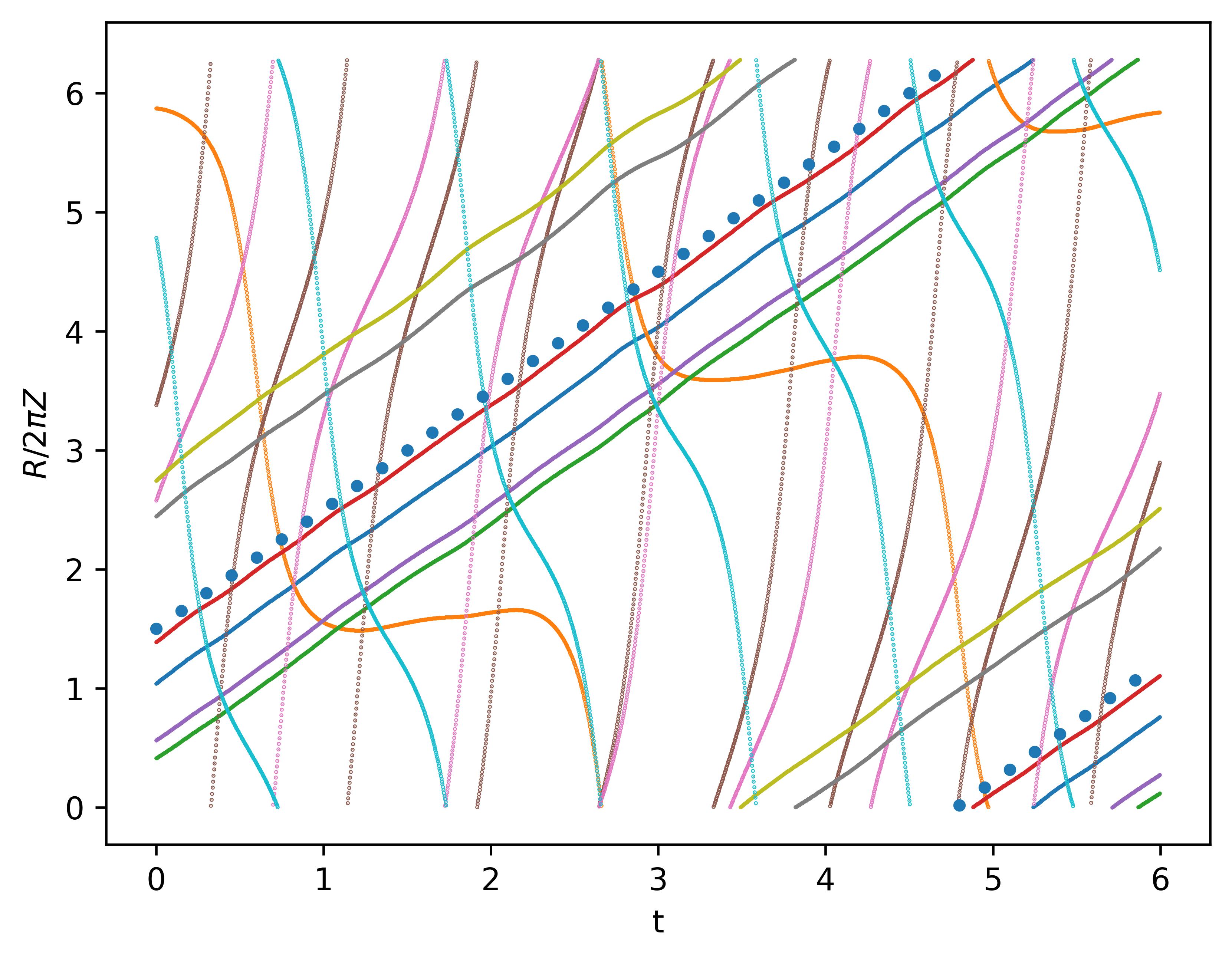}
    \caption{$F=6$ }
    \end{subfigure}
    \caption{Let $u: \R \rightarrow \T$ be given by $u(t)=c t$ with $c=1$, the blue dotted line in the figures, and $K=1$. Then, for $F$ sufficiently large (a) the states $\theta$ are entrained with $u$. For $F$ sufficiently small (b) solutions can break away from $u$. }  \label{fig:forcedkura}
\end{figure}

For sufficiently large $F$ the oscillators $\theta_k$ lock to $u$ (Fig. \ref{fig:forcedkura}a). As $F$ is decreased oscillators start breaking free from $u$ (Fig. \ref{fig:forcedkura}b).

\subsubsection{Forced Kuramoto equations oscillator network \label{sec:forced}}

The prediction task is given by the input $u(t) = ct$. We will formulate a procedure that does not require a training phase. More specifically, we present an explicit function for the prediction that is independent of $c$.

Relying on the techniques developed in \cite{childs2008stability,ott2008low,antonsen2008external} we can consider the continuum limit equations \eqref{eq:forcedkura}, $N \rightarrow \infty$, for the complex order parameter, $z := \frac{1}{N} \sum_{k=1}^N e^{i \theta_k}$, by the considering so-called Ott-Antonsen ansatz. We need to assume that $\omega_k$ are sampled from a Cauchy distribution \eqref{eq:cauchy}. We set $\Delta_0=1$ as it turns out we can scale away $\Delta_0$. Let $r,\Psi$ be given by $r e^{i \Psi} = z$ then the governing equations in terms of $r$ and $\Psi$ are given by
\begin{gather}
\begin{aligned}
\dot r &=  -  r + \frac{K}{2}r (1-r ^2) + \frac{F}{2}(1-r^2) \cos (\Psi- u) ,\\
\dot \Psi &=  \omega_0 - \frac{F}{2} \left( r + \frac{1}{r} \right) \sin (\Psi - u).
\end{aligned} \label{eq:rhopsi_ct}
\end{gather}
The above equations are used for the wipe-out phase. From the analysis in \cite{childs2008stability} it follows that the system corresponding to $r$ and $\phi := \Psi-ct$ has stable fixed points for certain parameters. Denote a stable fixed point by  $(r_0,\phi_0)$. As a prediction for $u$ we then take $\Psi - \phi_0$. Note that different from the framework in \ref{sec:frame} we now have that $\hat{u}(t):= \Psi(t) - \phi_0 $. Hence, there is no training phase. During testing, the equations will be given by
\begin{gather}
\begin{aligned}
\dot r &=  -  r + \frac{K}{2} r (1- r^2) + \frac{F}{2}(1-r^2) \cos (\phi_0), \\
\dot \Psi &= \omega_0 - \frac{F}{2} \left( r + \frac{1}{r} \right) \sin (\phi_0).
\end{aligned} 
\end{gather}
Consequently, if the $r$-equation has a stable fixed point at $r_0$ then for any $\delta>0$ and $T_{\rm test}>0$ there exists a $T_{\rm wipe}$ such that  $|\hat{u}(t) - u(t)| < \delta  $ for all $t \in [0,   T_{\rm test}]$.  A full analysis of the parameter space can be performed using classical bifurcation theory such as used in \cite{childs2008stability}. It is outside the scope of the Supplementary Information.

In Equation \eqref{eq:rhopsi_ct} the continuum limit is given for $M=1$. The techniques in \cite{ott2008low} can be extended to $M$-dimensional input. Each input dimension will generate one dependent complex variable in the continuum limit. Hence, if the chaotic input is generated by an autonomous first order ODE the analysis of the system would require the study of (at least) a 6-dimensional first order autonomous ODE.

\section{Supplementary Information to Subsection 2.5}

\subsection{Extracting chaos generating maps from the order parameter}

\begin{figure}[ht]
\centering
	\begin{subfigure}[b]{0.45\textwidth}
	\centering
	\includegraphics[width = 7cm]{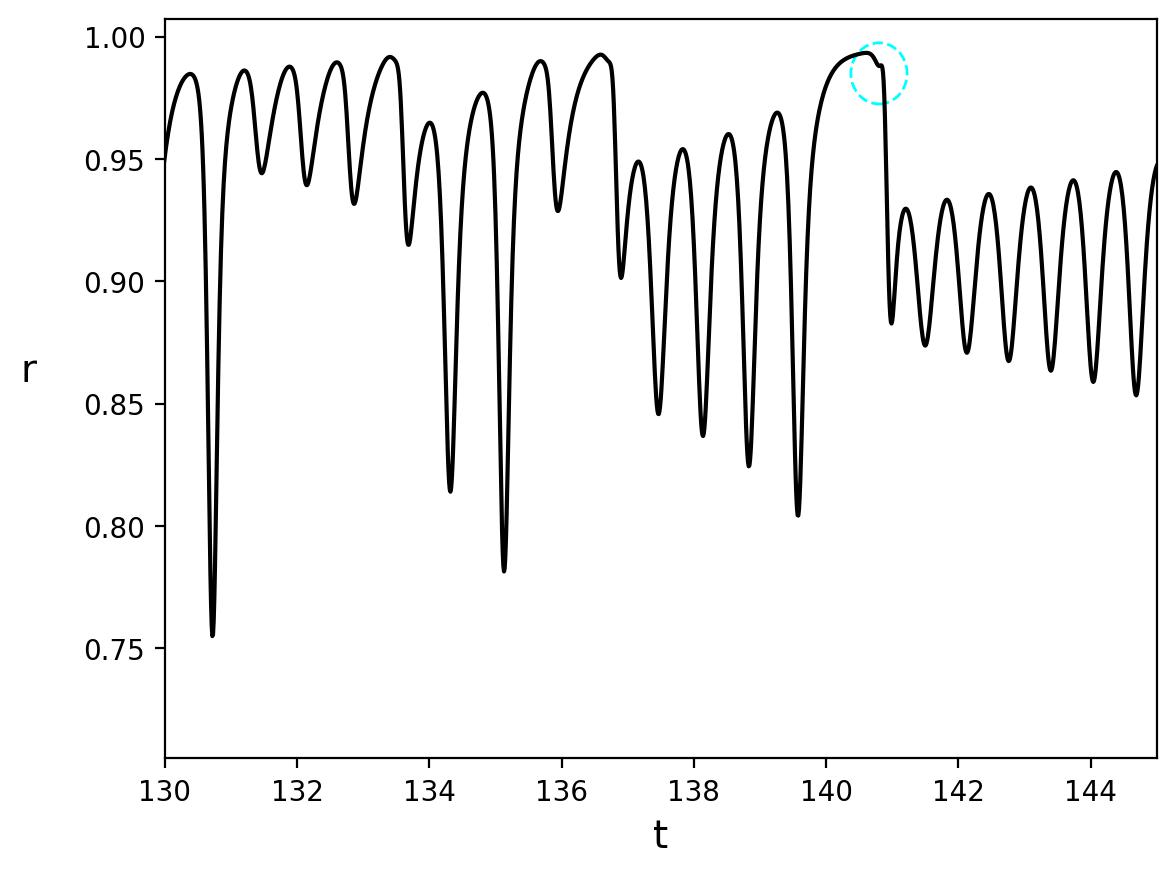}
        \caption{Time series $r$ during testing}
	\end{subfigure}
    \begin{subfigure}[b]{0.45\textwidth}
	\centering    
    \includegraphics[width= 7cm]{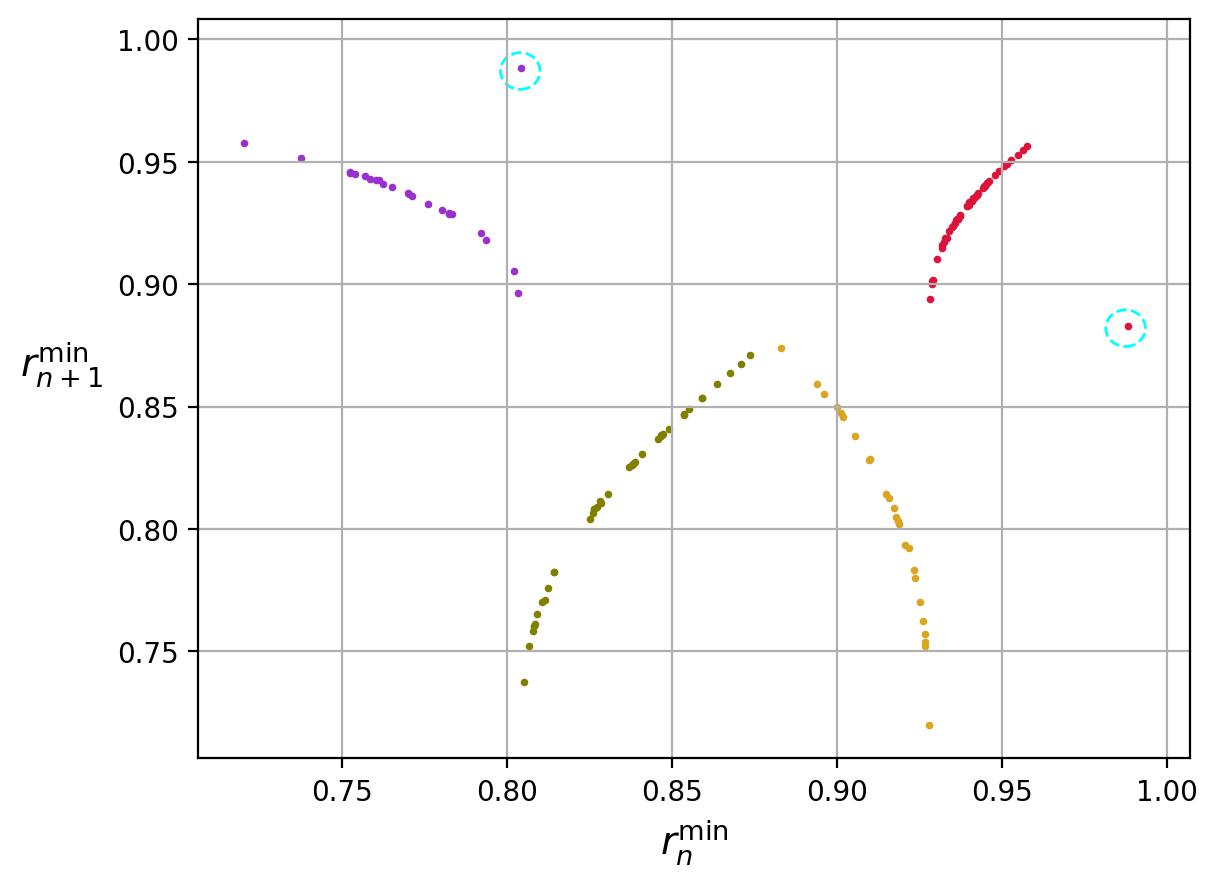}
    \caption{Graph of consecutive local minima of $r$}
    \end{subfigure}
    \caption{Anomaly in $r$: We consider $r$ for $t \in [100,200]$. In terms of local minima there occurs (a) a single local minima which has a small absolute second derivative in comparison to the other local minima,  circle in cyan. (b) The anomaly in $r$ translates to two anomalies in the graph corresponding to $r_n^{\min} \mapsto r_{n+1}^{\min}$ . }  \label{fig:rmin_anom}
\end{figure}

When computing the local minima in (Fig. 3c)  of the main paper we observed that small fluctuation in $r$ lead to local minima. These anomalies translate to anomalies when we graph consecutive local minima (Fig. \ref{fig:rmin_anom}). However, these can be easily identified by putting bounds on the second derivative which is done to obtain the figure in the main paper.

\begin{figure}[ht]
\centering
	\begin{subfigure}[b]{0.45\textwidth}
	\centering
	\includegraphics[width = 6.8cm]{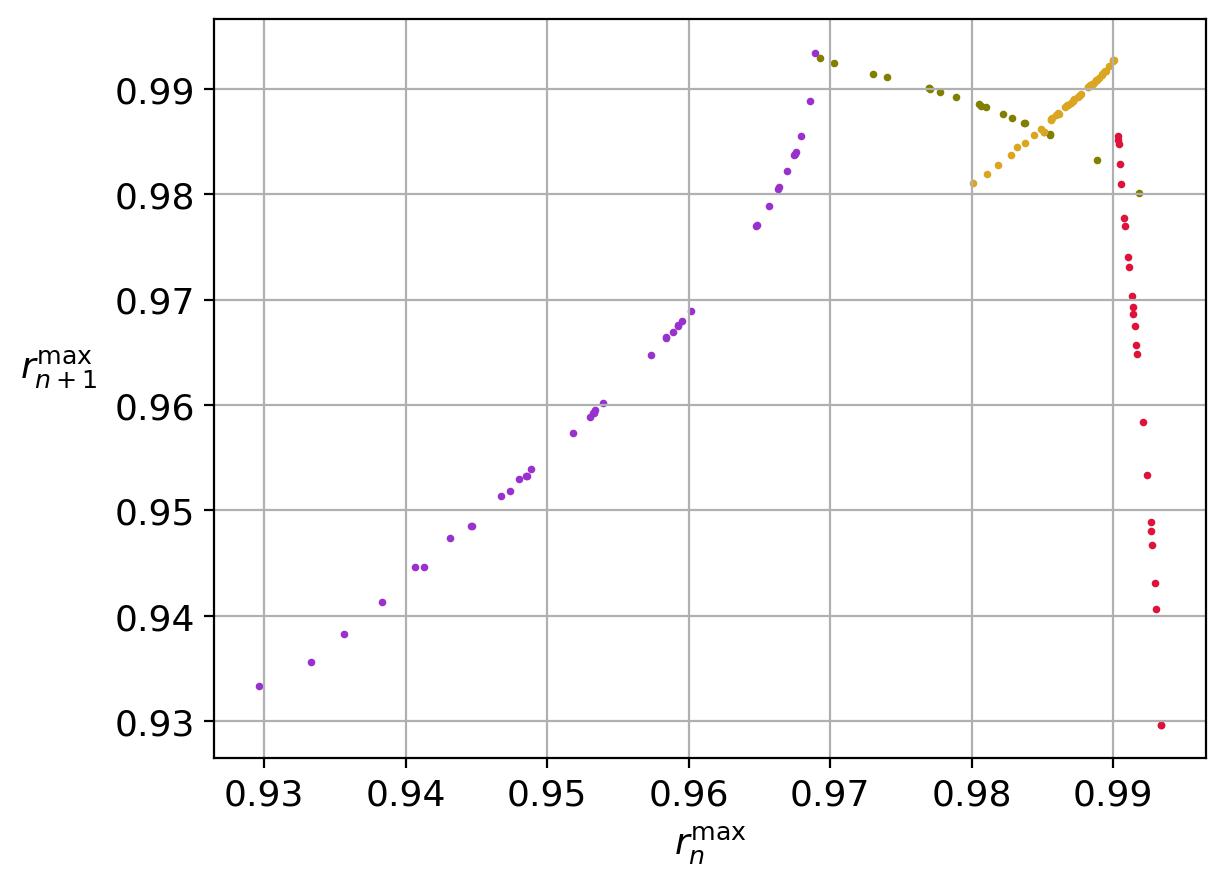}
        \caption{Graph of consecutive local maxima of $r$}
	\end{subfigure}
    \begin{subfigure}[b]{0.45\textwidth}
	\centering    
    \includegraphics[width= 6.5cm]{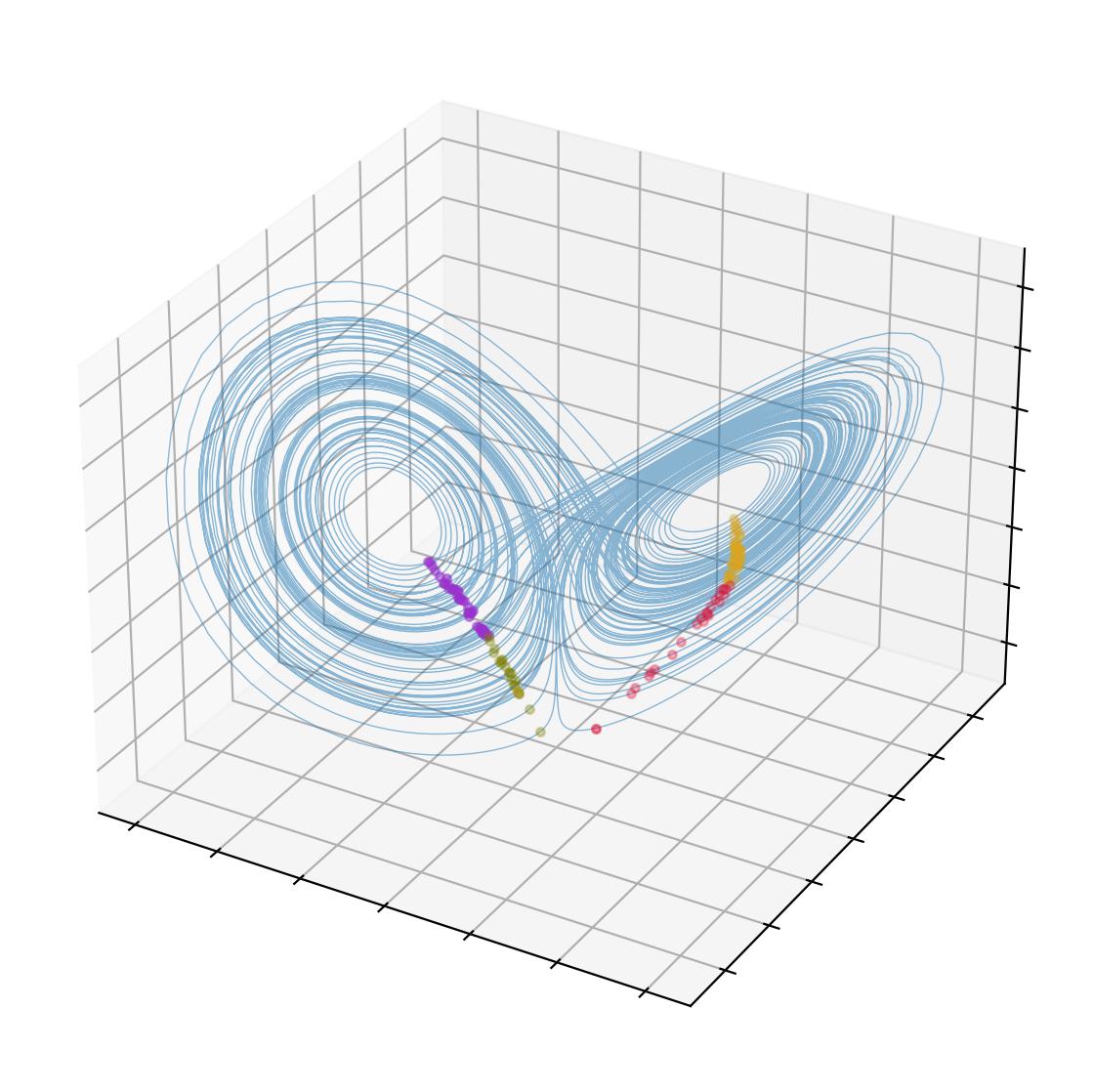}
    \caption{Attractor with local maxima domains of $r$}
    \end{subfigure}
    \caption{Dynamics of consecutive local maxima of $r$: By considering the map $r_n^{\max} \mapsto r_{n+1}^{\max}$ we can find a graph that describes the chaotic dynamics of the attractor. Four domains can be identified which on the attractor (b) corresponds to sections which have the characteristic dynamics of the Lorenz system. }  \label{fig:rmax}
\end{figure}

Instead of considering local consecutive minima we can also perform the procedure for consecutive maxima of $r$. Consider the map $r_n^{\max} \mapsto r_{n+1}^{\max}$. Certain domains for the symbolic mapping are disconnected contrary to the results in the figure of the main paper  (Fig. \ref{fig:rmax}a). However, the domains correspond to the same symbolic and attractor dynamics where after an iteration there is a domain that extends to its neighbour on the same wing and a domain that moves to the other wing and covers the full wing (Fig. \ref{fig:rmax}b).  

We note that the local maxima of $r$ cover a much smaller range of $r$ compared to local minima of $r$. Since we wanted to visualize the projection of the critical points on the range of $r$ it was easier to use the local minima. 

\subsection{Dependence of local minima of $r$ on $K$ \label{sec:rminK}}

\begin{figure}[ht]
\centering
\includegraphics[width=8cm]{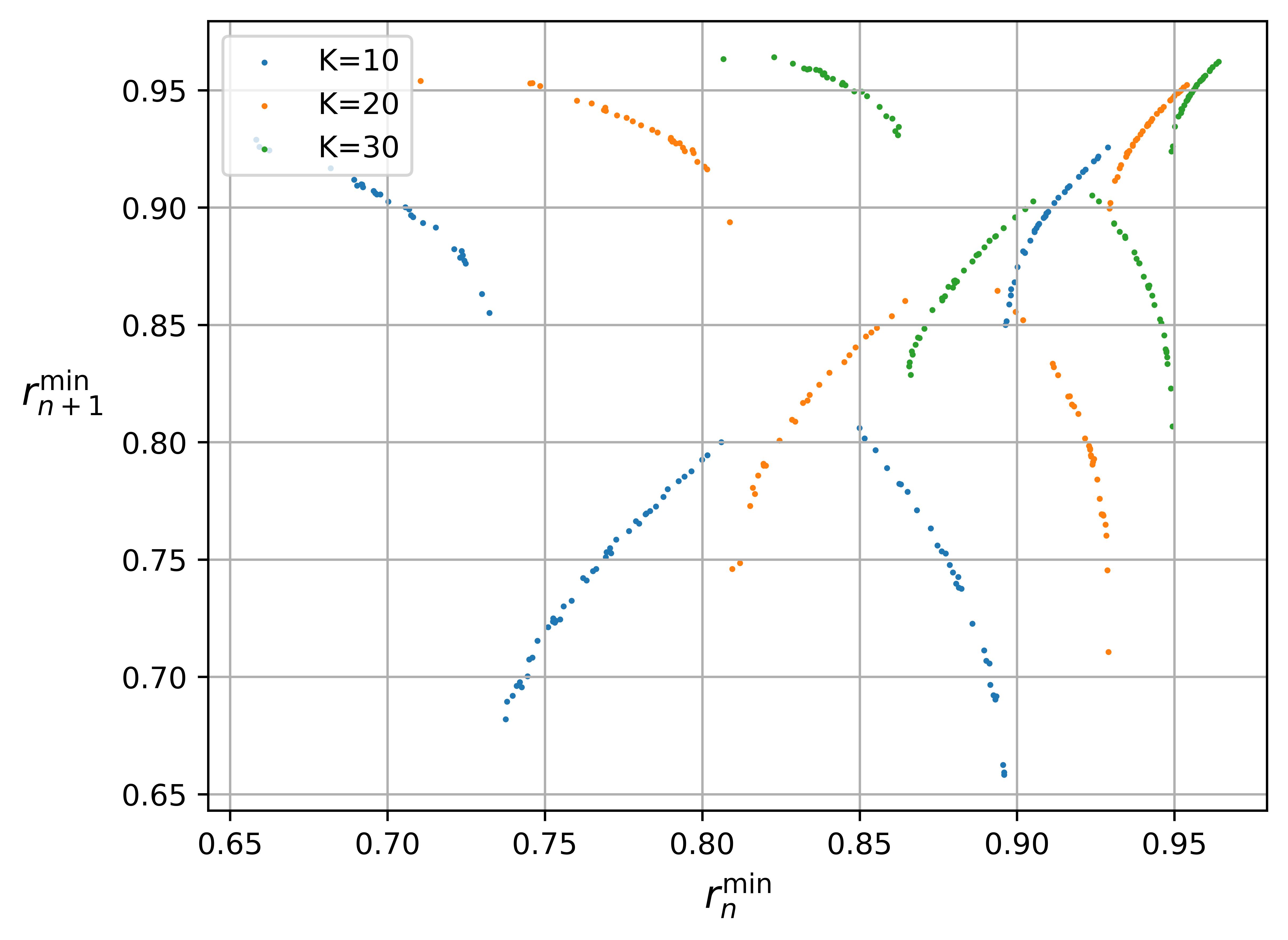}
    \caption{Graph of the map $r^{\min}_n \mapsto r^{\min}_{n+1}$ for varying $K$ and $F=35$.}
     \label{fig:rmin_K}
\end{figure}

Recall the chaotic map considered in the main paper given by $r^{\min}_n \mapsto r^{\min}_{n+1}$. Increasing $K$ leads to a decrease in the range of $r^{\min}$ (Fig. \ref{fig:rmin_K}). In particular, the global minimum of $r$ appears to be increasing with respect to $K$.  

\section{Supplementary Information to Subsection 2.6 \label{sec:lyabifu}}

\subsection{Lyapunov exponents}

\subsubsection{Jacobian}
The structure of the Kuramoto model allows us to efficiently evaluate the Jacobian. We define $\bar{R}_1(\theta) := \frac{1}{N} \sum_{j=1}^N \sin( \theta_j)$, $\bar{R}_2(\theta) :=  \frac{1}{N} \sum_{j=1}^N \cos( \theta_j)$. We write the governing equations \eqref{eq:gov} as
\begin{gather}
\begin{aligned}
\frac{d\theta_i }{dt} &= \omega_i + K \left[ \bar{R}_1(\theta) \cos(\theta_i) - \bar{R}_2(\theta) \sin( \theta_i)\right] + F \sin ( c \hat{u}_{v_i}(\theta)-\theta_i ).
\end{aligned}
\end{gather}
Denote the Jacobian by $J$. For the autonomous oscillator network we have that $\hat{u}(\theta)= W' [1, \sin(\theta), \sin^2(\theta) ]^{\top}$, $W' \in \R^{M \times (2N+1)}$. We decompose $W'$ as $W' = [W^b, W^1,W^2]$ with $W^b \in \R^{M \times 1}$, $W^1 \in \R^{M \times N}$, $W^2 \in \R^{M \times N}$. Denote the entries of $J$ by $b_{i,j}$ then we have that

\begin{align*}
b_{i,j} &= a_{i,j}  + c F  \cos(c \hat{u}_{\nu_i}(\theta) - \theta_i)    \left(   \frac{\partial \hat{u}_{v_i}(\theta)}{\partial \theta_j} \right) ,
\end{align*}
with 
\begin{align*}
a_{i,j} =   \frac{K}{N} \cos(\theta_j - \theta_i ) - \delta_{ij} \left( F \cos( \hat{u}_{v_i}(\theta) - \theta_i) +  \frac{K}{N} \sum_{k=1}^N \cos( \theta_k - \theta_i )    \right),
\end{align*}
and 
\begin{align*}
 \frac{\partial (\hat{u})_k(\theta)}{\partial \theta_j}  &= (W^1)_{k,j} \cos(\theta_j) + 2 (W^2)_{k,j} \cos(\theta_j) \sin(\theta_j),
\end{align*}
for $k=1,2,3, \ldots, M$. Note that $a_{i,j}=a_{j,i}$.

\subsubsection{Computation method}

The Lyapunov exponents are computed by using the Gram-Schmidt procedure in \cite{parker2012practical,benettin1980lyapunov}. We apply orthonormalization every other step.  We used this procedure to compute the largest three Lyapunov exponents of oscillator network. Convergence requires anywhere between 5 to 12 times the training length. This procedure is relatively computationally expensive as evolution of the variational equations has quadratic complexity. The parameter domain in the main paper requires about one week on one CPU. However, this processes can easily be parallelized if you have multiple CPUs available. Additionally, quadratic processes benefit from deployment on GPU. 

The fluctuations make it computationally expensive to consider higher order precision. Instead it is better to add a regression step. We can then find parameters such that the leading Lyapunov exponent of the oscillators and Lorenz system agree up to three decimals, e.g., $F=29.719, K=23.544$.

\subsubsection{Condition attractor climate}

\begin{figure}[ht]
    \centering
	\includegraphics[width = 7cm]{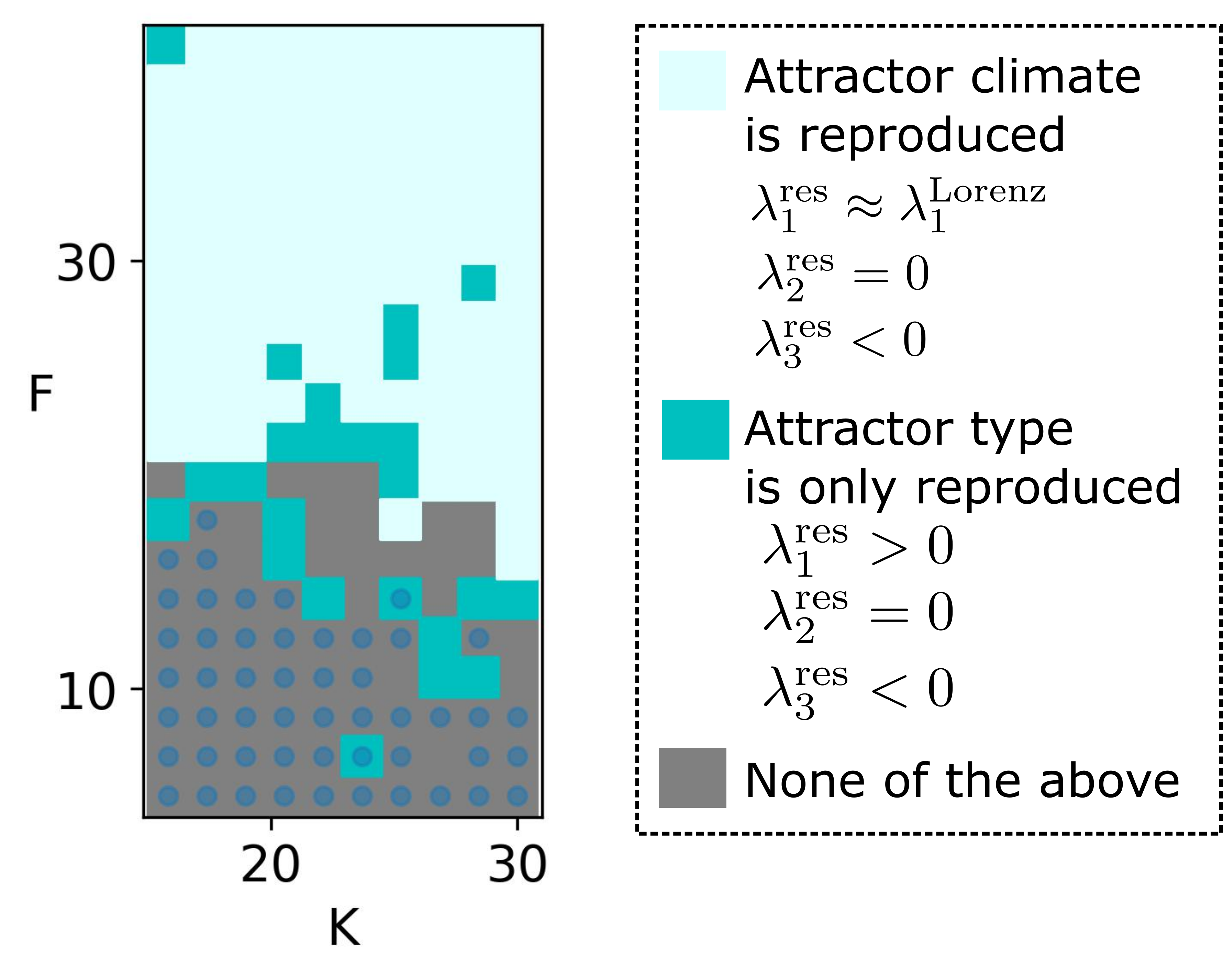}
        \caption{Attractor climate reproduction bifurcation diagrams: Numerical results for $\epsilon = 0.01$ } \label{fig:bif_esp01}
\end{figure}

Denote the Lyapunov exponents of the Lorenz attractor by $\lambda_j^{\rm Lorenz}$ with $j=1,2,3$ and let them be arranged in decreasing order. Denote the Lyapunov exponents of the oscillators by $\lambda_j^{\rm res}$ with $j=1,2, \cdots, N$ and let them also be arranged in decreasing order. Then, we say that the Lorenz attractor climate is reproduced when $\lambda_i^{\rm Lorenz} = \lambda_i^{\rm res}$ for $i=1,2$ and $\lambda_k^{\rm res}<0$ for $k>2$ \cite{lu2018attractor}.  The numerical condition becomes $|\lambda_i^{\rm Lorenz}- \lambda_i^{\rm res}| \leq \epsilon$ for $i=1,2$ and $\lambda_{k}^{\rm res}\leq  - \epsilon$  for $k>2$, where $\epsilon$ us a small positive constant.  We say that the attractor type is reproduced when $\lambda_1^{\rm res}>0$, $\lambda_2^{\rm res}=0$, $\lambda_k^{\rm res}<0$ for $k>2$. The numerical condition becomes $\lambda_1^{\rm res}>\epsilon$, $|\lambda_2^{\rm res}|<\epsilon$, $\lambda_k^{\rm res}<- \epsilon$. In (Fig. 4b) of the main paper we consider $\epsilon = 0.02$. However, the results for $\epsilon =0.01$ only differ slightly (Fig. \ref{fig:bif_esp01}).

\subsubsection{Lyapunov bifurcation diagram}

\begin{figure}[ht]
\centering
    \centering
\includegraphics[width=6cm]{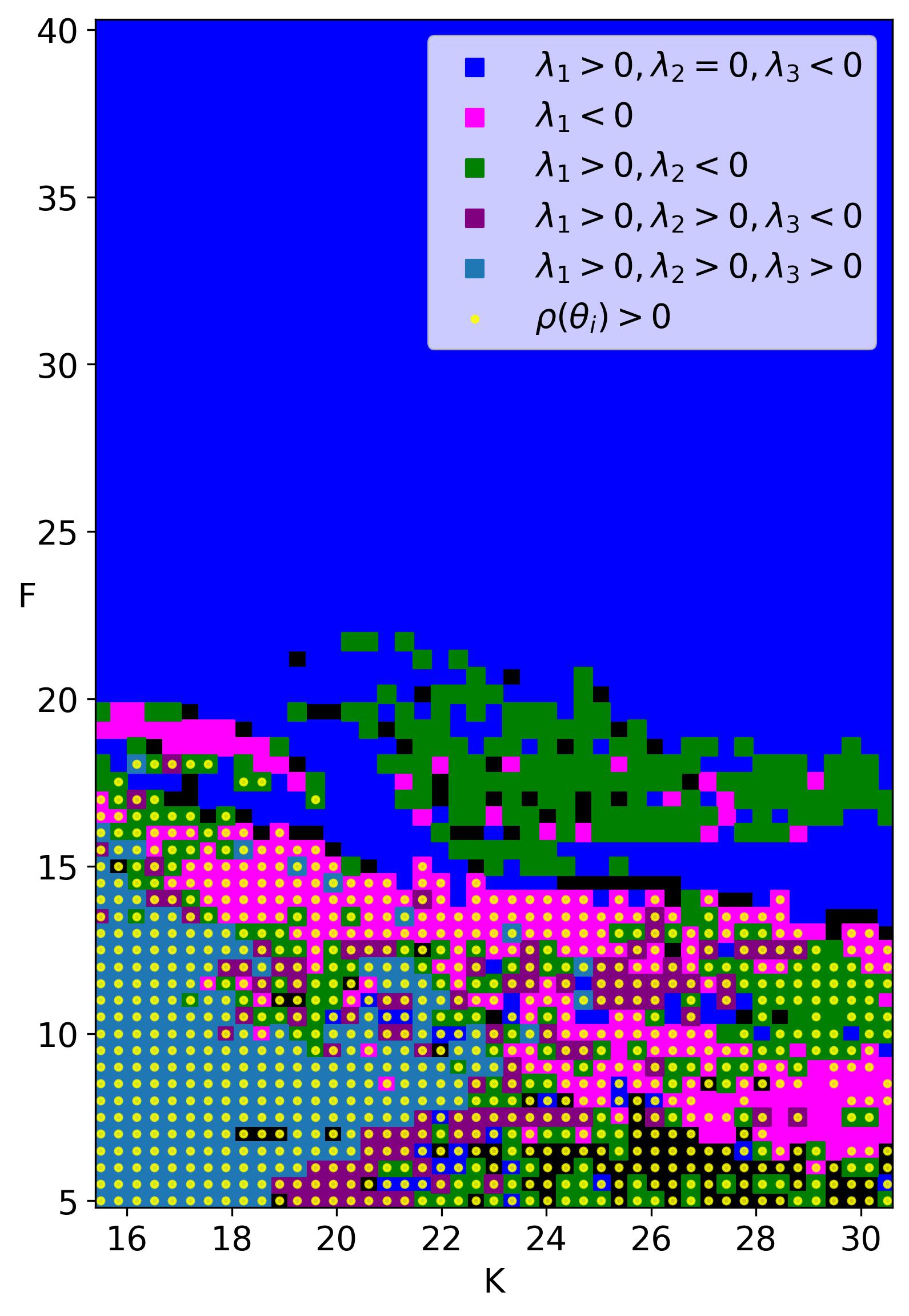}
    \caption{Bifurcation for the leading three Lyapunov exponents: The blue domain corresponds to chaotic attractors which belong to the same family as the Lorenz attractor. The black domain requires higher number of iterations to determine the sign of all the Lyapunov exponents. }  \label{fig:full_lya}
\end{figure}

In (Fig. \ref{fig:full_lya}) we consider the sign of the leading three Lyapunov exponents of the oscillators to give more detailed information about the dynamics.

Note that because of the compactness of $\T^N$ we do not have that at least one Lyapunov exponent vanishes if the trajectory of an attractor does not contain a fixed point \cite{haken1983least}.

\subsection{Rotation number}

\subsubsection{Numerics}
To evaluate the rotation number, $\rho(\theta_k)$, we evaluate the autonomous oscillator network for $2 \cdot 10^4$ time-steps which is equivalent to double of the training length.

\subsubsection{Effect of rotations in the states on the order parameter}

\begin{figure}[hbt!]
\centering
	\begin{subfigure}[b]{0.45\textwidth}
	\centering
	\includegraphics[width = 6.2cm]{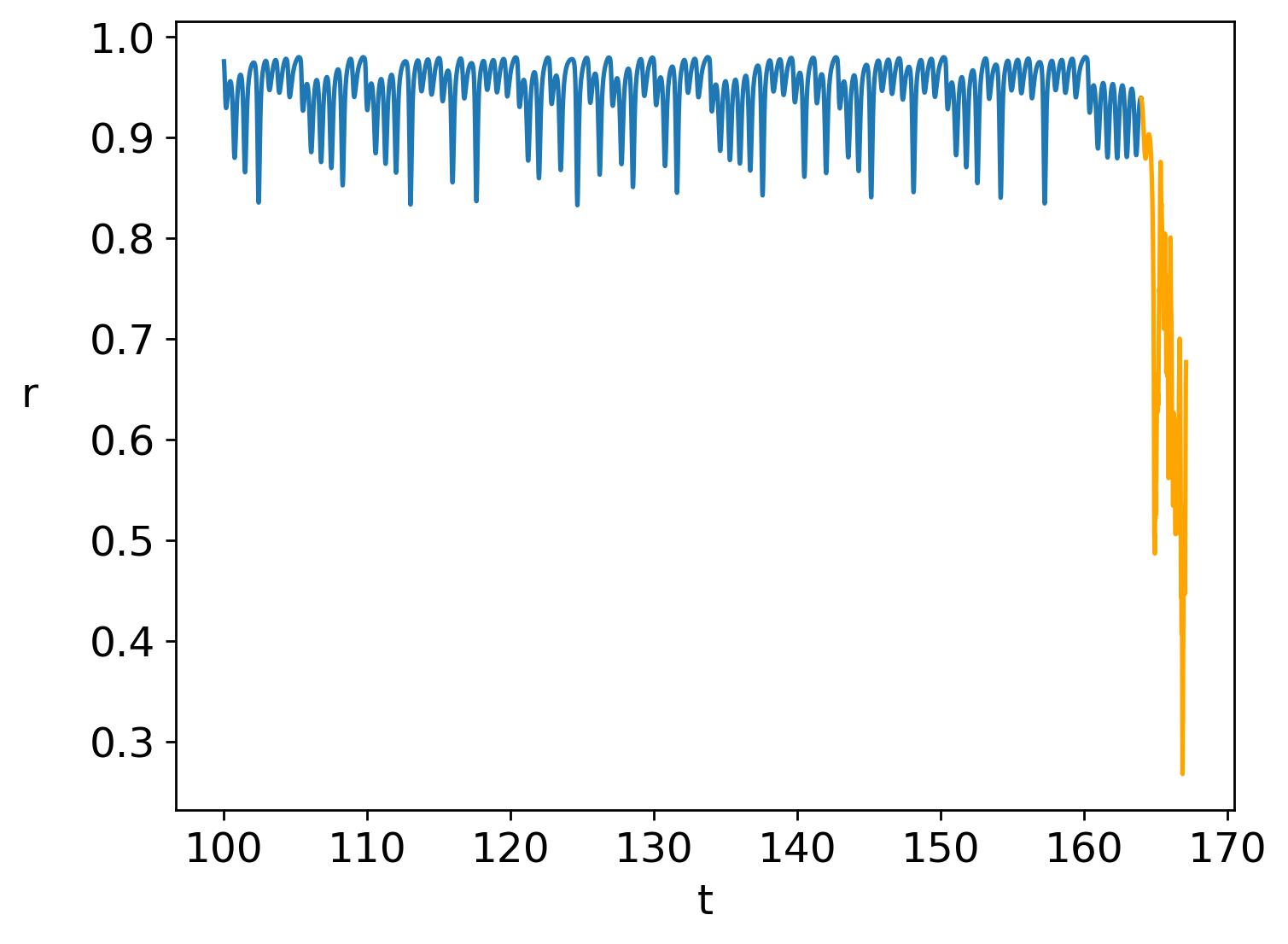}
        \caption{Evolution of $r$}
	\end{subfigure}
    \begin{subfigure}[b]{0.45\textwidth}
	\centering    
    \includegraphics[width= 6.2cm]{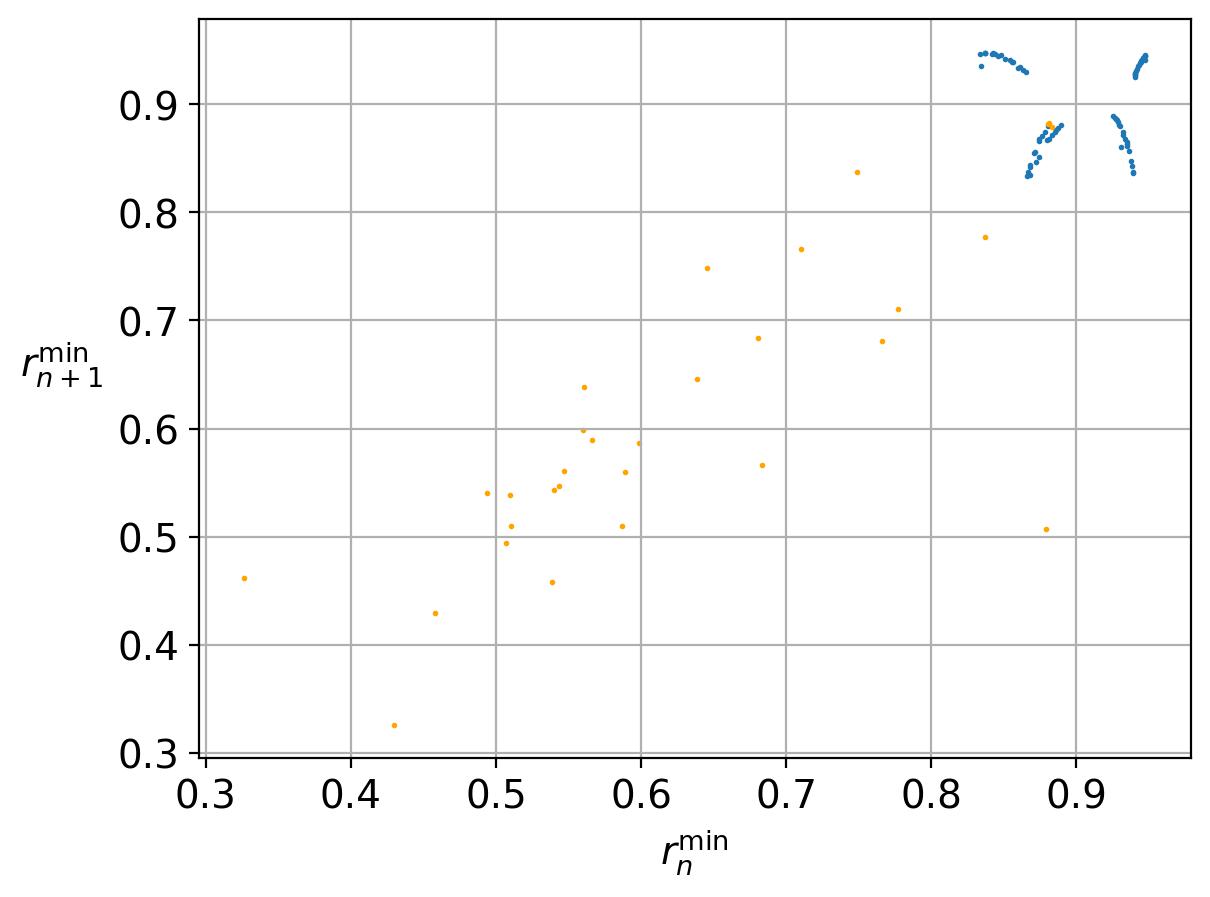}
    \caption{Dynamics of the map $r^{\rm min}_{n} \mapsto r^{\rm min}_{n+1}$ }
    \end{subfigure}
    \caption{Collapse of the learned dynamics in $r$: We consider the setting in (Fig. 4d) of the main paper with the same color coding. The $r$-dynamics collapses which translates to a collapse in the learned chaotic dynamics. } \label{fig:rcollapse}
\end{figure}

In (Fig. \ref{fig:rcollapse}) we investigate the order parameter $r$ for the case considered in (Fig. 5d) of the main paper. 
 We observe  in (Fig. \ref{fig:rcollapse}a) that $r$ suddenly decreases as full rotations occur. Furthermore, the collectively learned dynamics encoded in the map $r^{\rm min}_{n} \mapsto r^{\rm min}_{n+1}$ also gets lost when the states start making full rotations (Fig. \ref{fig:rcollapse}b). 

\subsubsection{Continuum limit for constant input}

In the continuum limit the average rotations that the oscillators make will be connected to $\Psi$ defined in Section \ref{sec:forced}. Let us consider a constant $u$. Then, for suitable parameters Equation \eqref{eq:rhopsi_ct} can have a stable limit cycle arising from a Hopf bifurcation~\cite{childs2008stability}. These stable limit cycles in a finite-dimensional setting would lead to a diverging rotation number. Note that the rotation number can only diverge for $t \rightarrow \infty$ since the vector field \eqref{eq:gov} is bounded.

\section{Music box task \label{sec:music}}

The contents of this section concerns the technique behind (Fig. 1b-iii). The task is to memorize and produce sound from a looped piece of sheet music. The sheet music is learned by two Kuramoto oscillator networks and for each pitch in the sheet music a oscillator network is trained that produces the soundwave for that specific pitch. The evolution of the oscillator networks is visualized in the accompanying music videos. \\

\textbf{Melody I}: (Fig.1-b-iii): Kuramoto oscillator networks are stacked to memorize melodies\\

\textbf{Melody II}: (Fig.1-b-iii): Kuramoto oscillator networks are stacked to memorize melodies

\subsection{Learning sheet music}

The notes in the sheet music are separated into pitches with a corresponding pitch duration which we refer to as beat. We store these into two separate lists. Below we wrote the lists corresponding to the first bar of the sheet music in (Fig. 1b-iii) of the main paper:

\begin{align*}
\text{ pitch list} =  \left[\text{C4,\,D4,\,F4,\,D4,\,A4,\,A4,\,G4,\,} \ldots \right], \; \;  
\text{ beat list} = \left[\text{1/16,\,1/16,\,1/16,\,1/16,\,3/16,\,3/16,\,6/16,\,} \ldots \right] .
\end{align*}

\begin{figure}[hbt!]
\centering
	\begin{subfigure}[b]{0.45\textwidth}
	\centering
	\includegraphics[width = 6.3cm]{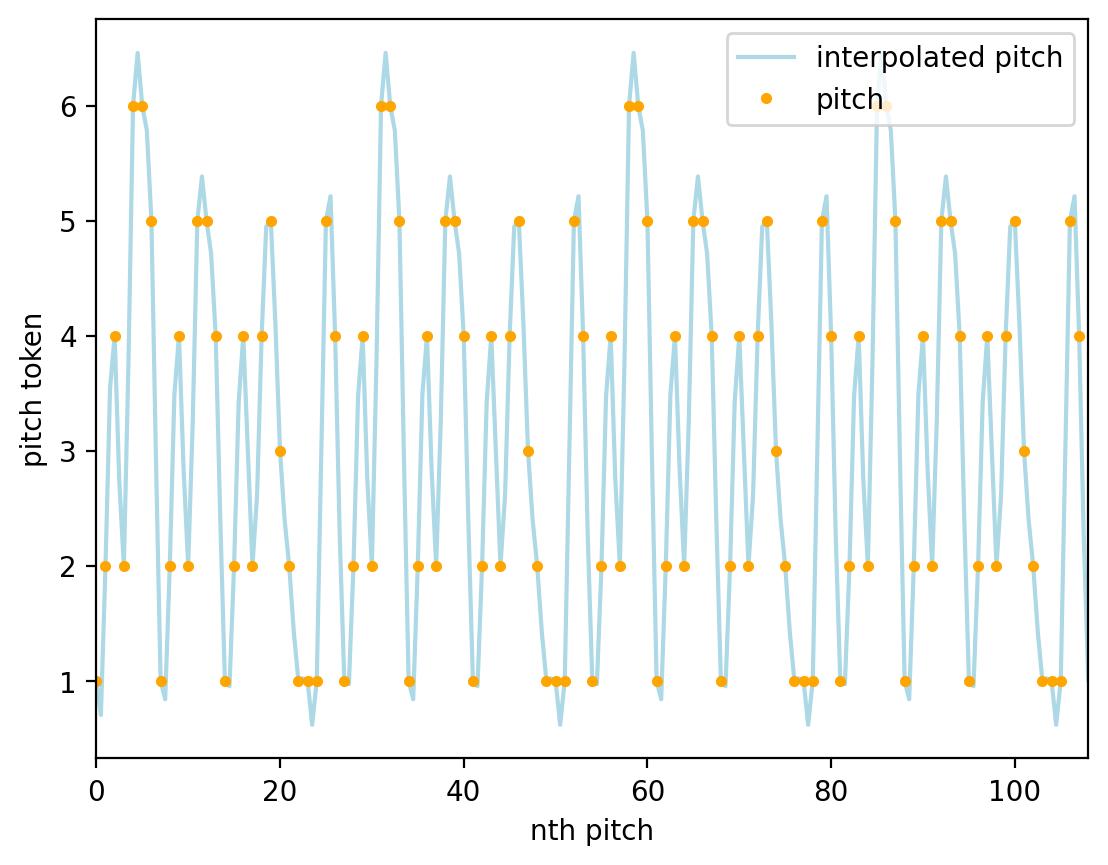}
        \caption{Pitch}
	\end{subfigure}
    \begin{subfigure}[b]{0.45\textwidth}
	\centering    
    \includegraphics[width= 6.3cm]{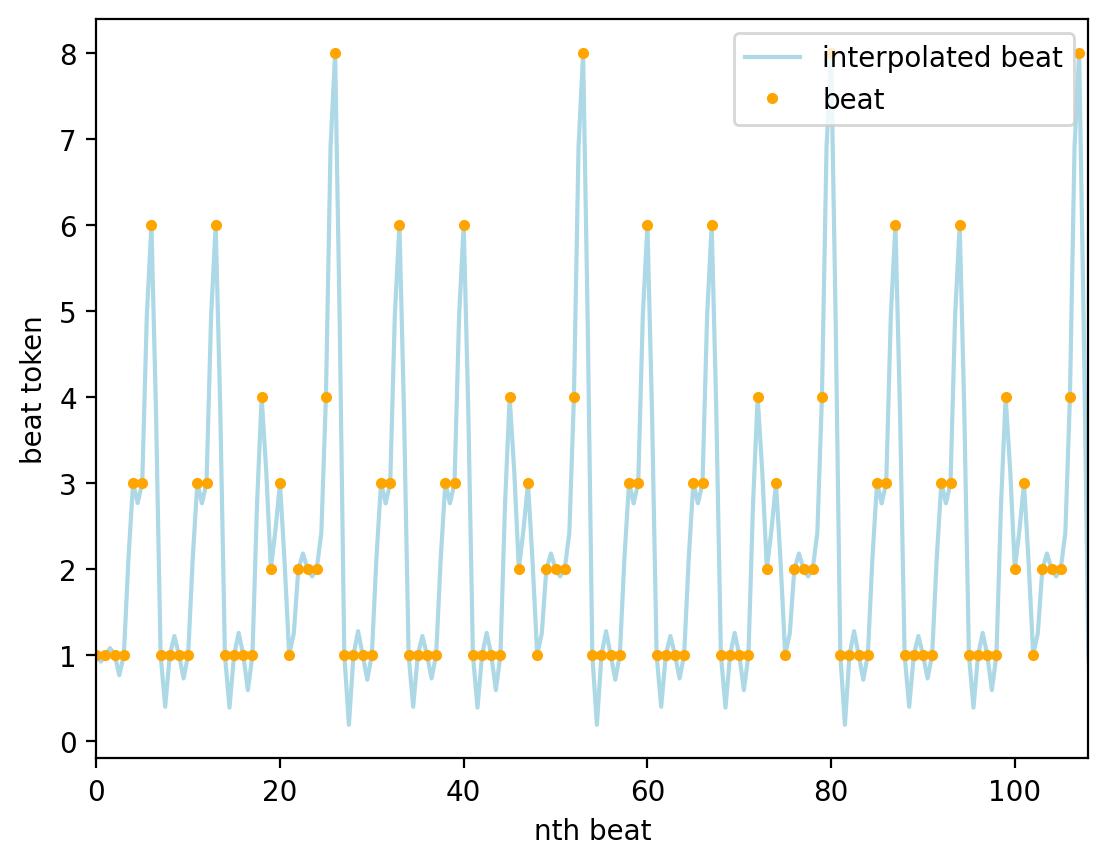}
    \caption{Beat}
    \end{subfigure}
    \caption{From discrete to continuous input: after mapping pitches and beats to  integers we interpolate the discrete data using cubic splines to arrive at a continuous input on which we can train the oscillator networks.}
    \label{fig:disc_to_cont_sheet}
\end{figure}

We then stack these lists to arrive at a periodic sequence. Since the Kuramoto oscillator network is a continuous system it is difficult to train it on a discrete input. Hence, we first map the values in the lists to unique integers and then apply cubic splines to obtain an interpolation that is used as continuous input for the network (Fig. \ref{fig:disc_to_cont_sheet}). These periodic continuous inputs are then learned by two separate oscillator networks. To retrieve the pitches or beats we sample at the frequency corresponding to the discrete input. 

\subsection{From pitches to sound}

\begin{figure}[ht]
    \centering
    \includegraphics[width=9cm]{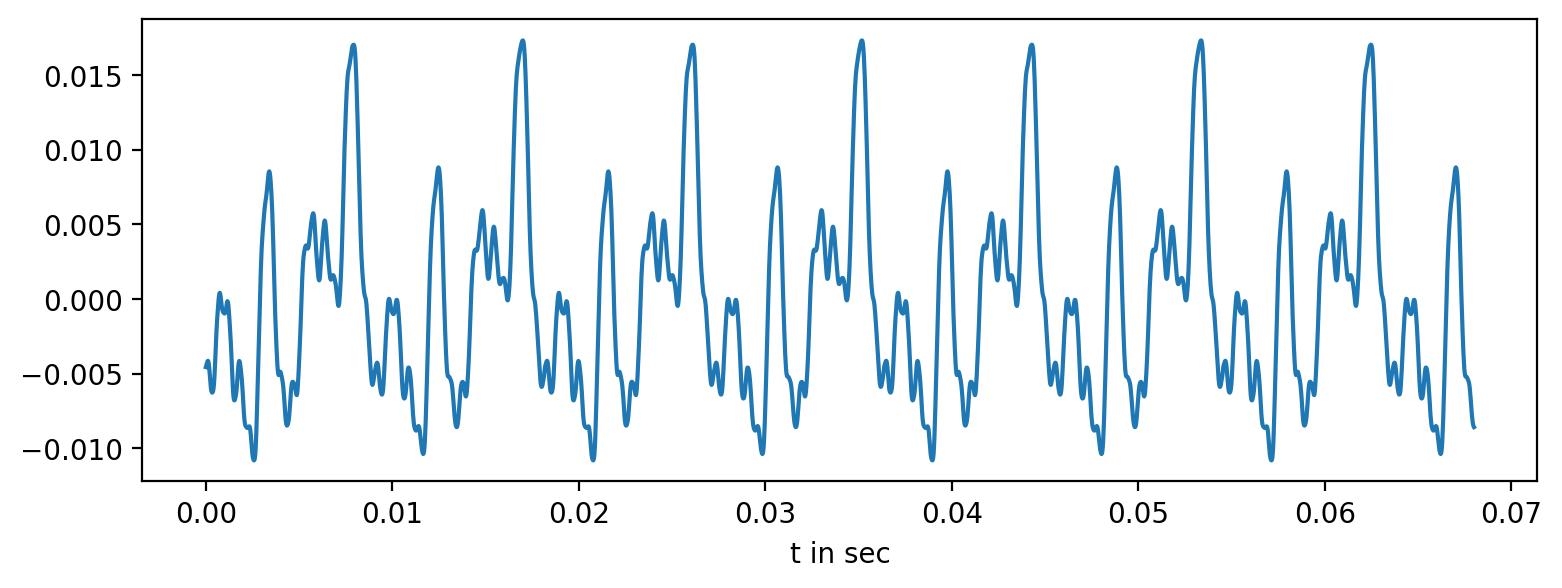}
    \caption{Sound profile of an A4 pitch for a reed organ.} 
    \label{fig:a4}
\end{figure}

For each pitch in the pitch list we train a sound oscillator network which is learned from an instrument soundfile sampled at 22050Hz. Smoothness of the input sound profile is important for the Kuramoto oscillator network to learn the input. To see this, observe that if the input is continuous then the vector field corresponding to the Kuramoto oscillator network is continuous and hence the prediction must be $C^1$-smooth. Organs have very smooth sound profiles. Here, we sampled pitches from a reed organ. A sound profile for an A4 pitch is displayed in (Fig. \ref{fig:a4}).  

\subsection{Connecting sheet music oscillator networks to sound oscillator networks}

\begin{figure}[hbt!]
    \centering
     \includegraphics[width=8cm]{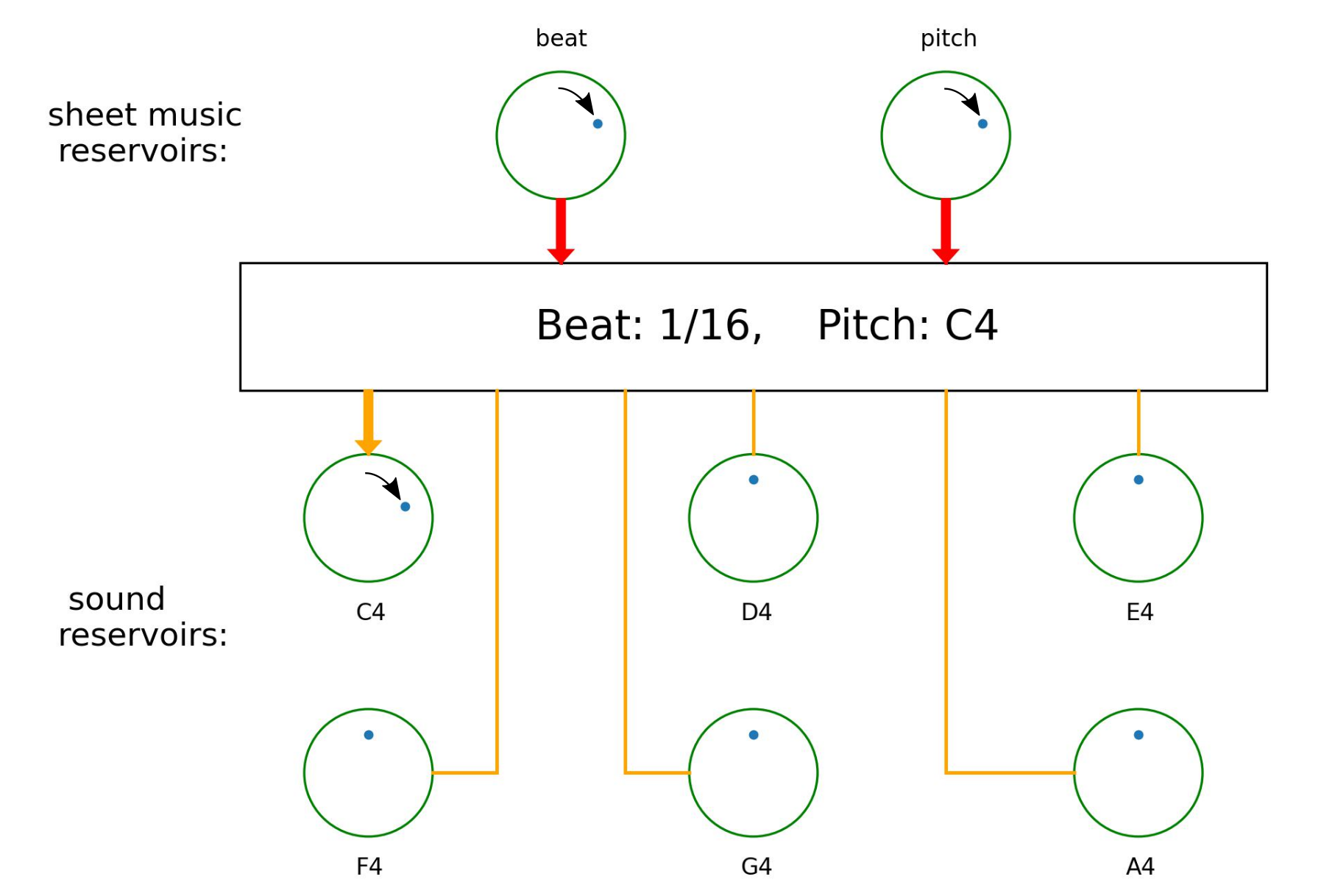}
    \caption{Architecture of the music oscillator networks: The beat and pitch oscillator networks are evolved to produce a beat and a pitch which initialize the corresponding sound producing pitch oscillator networks for the duration of the outputted beat.}
    \label{fig:music_arch}
\end{figure}

We now put the trained oscillator networks together (Fig. \ref{fig:music_arch}). 

We evolve the pitch and beat oscillator networks which output a pitch with corresponding beat. The outputted pitch initiates the evolution of the corresponding sound producing pitch oscillator network for the duration of the outputted beat. The architecture is animated in the accompanying videos.

\section{R\"{o}ssler system \label{sec:rossler}}

The R\"{o}ssler system~\cite{rossler1976equation} is given by the ordinary differential equation
\begin{align*}
    \frac{dx}{dt} &= -y -z ,\\
    \frac{dy}{dt} &= x+ a y, \\
    \frac{dz}{dt} &= b + z(x-c).
\end{align*}
We consider the parameters $(a,b,c)=( 0.2, 0.2, 5.7)$ for which the system has a chaotic attractor. 

\begin{figure}[hbt!]
    \centering
    \includegraphics[width=16cm]{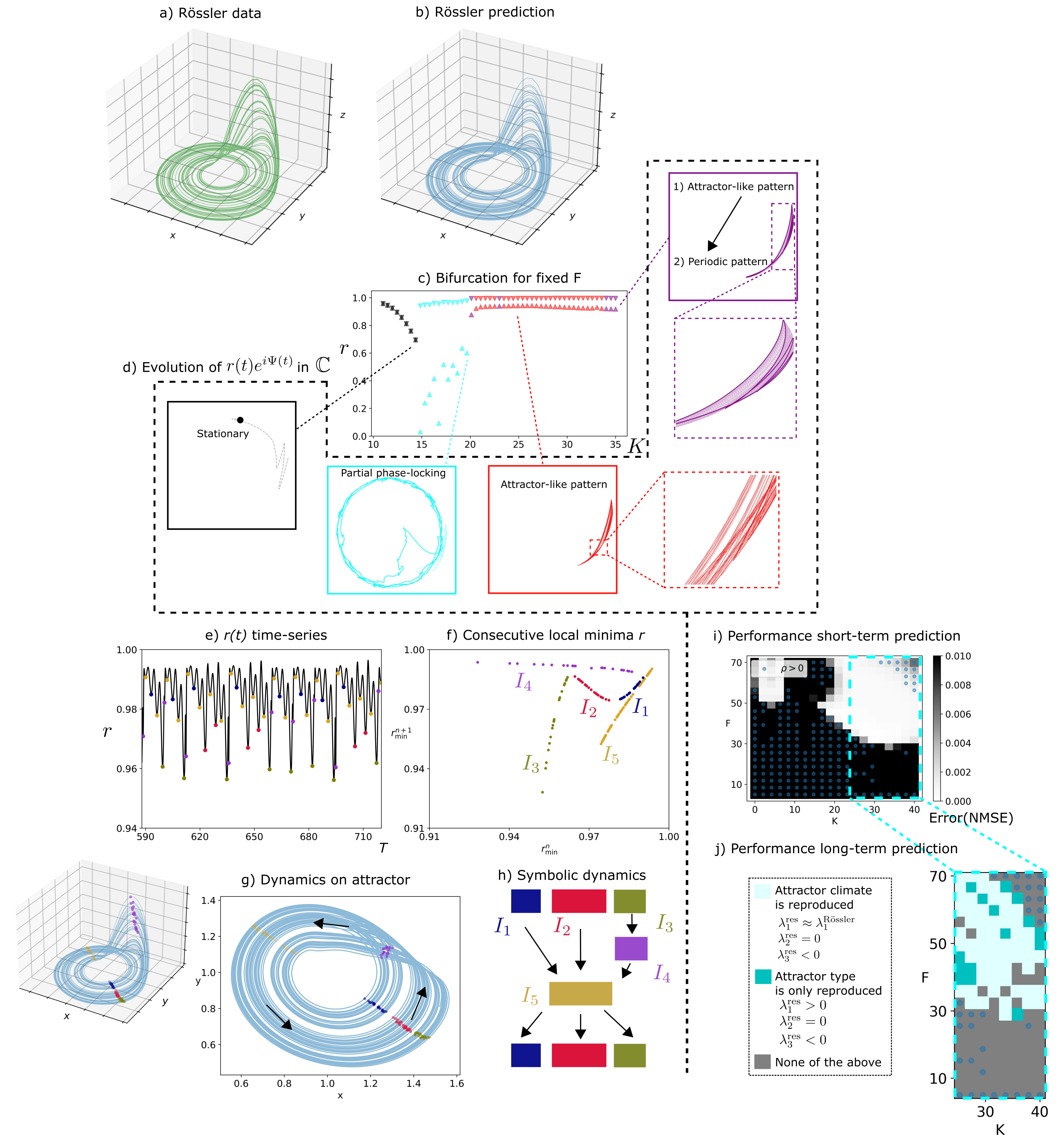}
    \caption{Experiments of the main paper for input time-series on the R\"{o}ssler attractor.}
    \label{fig:roes}
\end{figure}

We repeat the analysis from the main paper for the R\"{o}ssler attractor (Fig. \ref{fig:roes}). We observe that the data and the prediction resemble each other in the state space (Fig. \ref{fig:roes}ab).  We investigate the bifurcation for fixed $F$ and varying $K$ using the order parameter $r$ (Fig. \ref{fig:roes}c). This bifurcation is different from the Lorenz case discussed in the main paper. However, it appears to resemble the bifurcation occurring for Lorenz with fixed $K$ and varying $F$ from Section \ref{sec:bifuF}. In the R\"{o}ssler system the $x,y$-components perform a type of rotation and the $z$-component exhibits spike-like evolution. Hence, the dynamics is different from the Lorenz system and the $x$, $y$- and $z$-dynamics are different. However, the difference in dynamics for $x,y$-components and $z$-component of the R\"{o}ssler can be subdued by applying a suitable rotation matrix to the dependent variables. We observe that for small $K$ the oscillator states become stationary (Fig. \ref{fig:roes}d). As $K$ is increased partial phase-locking occurs. As $K$ is further increased we arrive at an attractor-like pattern. If $K$ is taken too large periodic motion occurs for the complex order parameter. 

We explore the time-series corresponding to the order parameter $r$ (Fig. \ref{fig:roes}e). As with the Lorenz system we consider local minima. We visualize consecutive local minima in $r$ and observe that we can identify 5 domains (Fig. \ref{fig:roes}f). We observe that these domains correspond to a section on the attractor to which we can identify symbolic dynamics (Fig. \ref{fig:roes}gh).

 We consider the NMSE of short-term evolution of the prediction in the $K,F$-parameter space (Fig. \ref{fig:roes}i). When $F$ or $K$ is small the error during testing is large. Contrary to the case with the Lorenz system we observe that the error increases when both $F$ and $K$ exceed a threshold. We also include the rotation number $\rho$. We observe that generally the short-term error is low whenever $\rho=0$. We investigate if the R\"{o}ssler attractor's climate is reproduced by comparing the Lyapunov exponents of the oscillators to the Lyapunov exponents of the R\"{o}ssler attractor (Fig. \ref{fig:roes}j).  Above an $F$ the parameters for which the climate is reproduced are abundant. However, if $K$ is too large the prediction exhibits periodic behaviour and the climate is not reproduced.

We note that the leading Lyapunov exponent of the R\"{o}ssler attractor is much smaller than the leading Lyapunov exponent of the Lorenz attractor: $\lambda_1^{\rm Lorenz}\approx 0.91,  \; \lambda_1^{\rm Rossler} \approx 0.07$. Hence, we would expect that the embedded attractor for the R\"{o}ssler system is more parameter-sensitive  than  the embedded attractor for the Lorenz system.

\bibliographystyle{alpha}
\bibliography{my_bib.bib}{}

\newcommand{\etalchar}[1]{$^{#1}$}
\begin{thebibliography}{TRAA{\etalchar{+}}17}

\bibitem[AFG{\etalchar{+}}08]{antonsen2008external}
T.M. Antonsen, R.T. Faghih, M.~Girvan, E.~Ott, and J.~Platig.
\newblock External periodic driving of large systems of globally coupled phase
  oscillators.
\newblock {\em Chaos: An Interdisciplinary Journal of Nonlinear Science},
  18(3), 2008.

\bibitem[AP00]{atiya2000new}
A.F. Atiya and A.G. Parlos.
\newblock New results on recurrent network training: unifying the algorithms
  and accelerating convergence.
\newblock {\em IEEE transactions on neural networks}, 11(3):697--709, 2000.

\bibitem[BdJD{\etalchar{+}}24]{baltussen2024chemical}
M.G. Baltussen, T.J. de~Jong, Q.~Duez, W.E. Robinson, and W.T.S. Huck.
\newblock Chemical reservoir computation in a self-organizing reaction network.
\newblock {\em Nature}, 631(8021):549--555, 2024.

\bibitem[BGGS80]{benettin1980lyapunov}
G.~Benettin, L.~Galgani, A.~Giorgilli, and J.M. Strelcyn.
\newblock Lyapunov characteristic exponents for smooth dynamical systems and
  for hamiltonian systems; a method for computing all of them. part 1: Theory.
\newblock {\em Meccanica}, 15:9--20, 1980.

\bibitem[Bol98]{bollobas1998random}
B.~Bollob{\'a}s.
\newblock {\em Modern graph theory}.
\newblock Springer, 1998.

\bibitem[CS08]{childs2008stability}
L.M. Childs and S.H. Strogatz.
\newblock Stability diagram for the forced kuramoto model.
\newblock {\em Chaos: An Interdisciplinary Journal of Nonlinear Science},
  18(4):043128, 2008.

\bibitem[CTS24]{chiba2024reservoircomputingkuramotomodel}
H~Chiba, K.~Taniguchi, and T.~Sumi.
\newblock Reservoir computing with the kuramoto model.
\newblock {\em arXiv:2407.16172}, 2024.

\bibitem[dJAT{\etalchar{+}}23]{de2023mlncp}
T.G. de~Jong, N.~Akashi, T.~Taniguchi, H.~Notsu, and K.~Nakajima.
\newblock Virtual reservoir acceleration for cpu and gpu: Case study for
  coupled spin-torque oscillator reservoir.
\newblock In {\em NeurIPS: ML with New Compute Paradigms Workshop}. NeurIPS,
  2023.

\bibitem[FFN{\etalchar{+}}18]{furuta2018macromagnetic}
T.~Furuta, K.~Fujii, K.~Nakajima, S.~Tsunegi, H.~Kubota, Y.~Suzuki, and
  S.~Miwa.
\newblock Macromagnetic simulation for reservoir computing utilizing spin
  dynamics in magnetic tunnel junctions.
\newblock {\em Physical Review Applied}, 10(3):034063, 2018.

\bibitem[FMW16]{flovik2016describing}
V.~Flovik, F.~Macia, and E.~Wahlstr{\"o}m.
\newblock Describing synchronization and topological excitations in arrays of
  magnetic spin torque oscillators through the kuramoto model.
\newblock {\em Scientific reports}, 6(1):32528, 2016.

\bibitem[GBMB21]{garg2021kuramoto}
N.~Garg, S.V.H. Bhotla, P.K. Muduli, and D.~Bhowmik.
\newblock Kuramoto-model-based data classification using the synchronization
  dynamics of uniform-mode spin hall nano-oscillators.
\newblock {\em Neuromorphic Computing and Engineering}, 1(2):024005, 2021.

\bibitem[Hak83]{haken1983least}
H.~Haken.
\newblock At least one lyapunov exponent vanishes if the trajectory of an
  attractor does not contain a fixed point.
\newblock {\em Physics Letters A}, 94(2):71--72, 1983.

\bibitem[HK70]{hoerl1970ridge}
A.E. Hoerl and R.~W. Kennard.
\newblock Ridge regression: Biased estimation for nonorthogonal problems.
\newblock {\em Technometrics}, 12(1):55--67, 1970.

\bibitem[Jae01]{jaeger2001echo}
H.~Jaeger.
\newblock The "echo state" approach to analysing and training recurrent neural
  networks-with an erratum note.
\newblock {\em Bonn, Germany: German National Research Center for Information
  Technology GMD Technical Report}, 148(34):13, 2001.

\bibitem[Jae02]{Jaeger2002}
H.~Jaeger.
\newblock A tutorial on training recurrent neural networks, covering bppt,
  rtrl, ekf and the "echo state network" approach.
\newblock Technical report, International University Bremen, 2002.

\bibitem[JH04]{jaeger2004harnessing}
H.~Jaeger and H.~Haas.
\newblock Harnessing nonlinearity: Predicting chaotic systems and saving energy
  in wireless communication.
\newblock {\em Science}, 304(5667):78--80, 2004.

\bibitem[KT76]{kuramoto1976persistent}
Y.~Kuramoto and T.~Tsuzuki.
\newblock Persistent propagation of concentration waves in dissipative media
  far from thermal equilibrium.
\newblock {\em Progress of theoretical physics}, 55(2):356--369, 1976.

\bibitem[Kur75]{kuramoto1975self}
Y.~Kuramoto.
\newblock Self-entrainment of a population of coupled non-linear oscillators.
\newblock In {\em International Symposium on Mathematical Problems in
  Theoretical Physics: January 23--29, 1975, Kyoto University, Kyoto/Japan},
  pages 420--422. Springer, 1975.

\bibitem[Kur84]{kuramoto1984chemical}
Y.~Kuramoto.
\newblock {\em Chemical turbulence}.
\newblock Springer, 1984.

\bibitem[LBAJ23]{lizier2023analytic}
J.T. Lizier, F.~Bauer, F.M. Atay, and J.~Jost.
\newblock Analytic relationship of relative synchronizability to network
  structure and motifs.
\newblock {\em Proceedings of the National Academy of Sciences},
  120(37):e2303332120, 2023.

\bibitem[LHO18]{lu2018attractor}
Z.~Lu, B.R. Hunt, and E.~Ott.
\newblock Attractor reconstruction by machine learning.
\newblock {\em Chaos: An Interdisciplinary Journal of Nonlinear Science},
  28(6), 2018.

\bibitem[Lor63]{lorenz1963deterministic}
E.N. Lorenz.
\newblock Deterministic nonperiodic flow.
\newblock {\em Journal of atmospheric sciences}, 20(2):130--141, 1963.

\bibitem[LPS15]{lam2015numba}
S.K. Lam, A.~Pitrou, and S.~Seibert.
\newblock Numba: A llvm-based python jit compiler.
\newblock In {\em Proceedings of the Second Workshop on the LLVM Compiler
  Infrastructure in HPC}, pages 1--6, 2015.

\bibitem[LWSJ19]{lymburn2019reservoir}
T.~Lymburn, D.M. Walker, M.~Small, and T.~J{\"u}ngling.
\newblock The reservoir's perspective on generalized synchronization.
\newblock {\em Chaos: An Interdisciplinary Journal of Nonlinear Science},
  29(9), 2019.

\bibitem[MG77]{mackey1977oscillation}
M.C. Mackey and L.~Glass.
\newblock Oscillation and chaos in physiological control systems.
\newblock {\em Science}, 197(4300):287--289, 1977.

\bibitem[MLR{\etalchar{+}}19]{markovic2019reservoir}
D.~Markovi{\'c}, N.~Leroux, M.~Riou, F.~Abreu~Araujo, J.~Torrejon, D.~Querlioz,
  A.~Fukushima, S.~Yuasa, J.~Trastoy, P.~Bortolotti, et~al.
\newblock Dynamic effects on reservoir computing with a hopf oscillator.
\newblock {\em Applied Physics Letters}, 114(1):012409, 2019.

\bibitem[Nak20]{nakajima2020physical}
K.~Nakajima.
\newblock Physical reservoir computing: an introductory perspective.
\newblock {\em Japanese Journal of Applied Physics}, 59(6):060501, 2020.

\bibitem[NFN{\etalchar{+}}19]{nakajima2019boosting}
K.~Nakajima, K.~Fujii, M.~Negoro, K.~Mitarai, and M.~Kitagawa.
\newblock Boosting computational power through spatial multiplexing in quantum
  reservoir computing.
\newblock {\em Physical Review Applied}, 11(3):034021, 2019.

\bibitem[OA08]{ott2008low}
E.~Ott and T.M. Antonsen.
\newblock Low dimensional behavior of large systems of globally coupled
  oscillators.
\newblock {\em Chaos: An Interdisciplinary Journal of Nonlinear Science},
  18(3), 2008.

\bibitem[PC12]{parker2012practical}
T.S. Parker and L.~Chua.
\newblock {\em Practical numerical algorithms for chaotic systems}.
\newblock Springer Science \& Business Media, 2012.

\bibitem[PJSF04]{peitgen2004chaos}
H.O. Peitgen, H.~J{\"u}rgens, D.~Saupe, and M.J. Feigenbaum.
\newblock {\em Chaos and fractals: new frontiers of science}, volume 106.
\newblock Springer, 2004.

\bibitem[PLH{\etalchar{+}}17]{pathak2017using}
J.~Pathak, Z.~Lu, B.R. Hunt, M.~Girvan, and E.~Ott.
\newblock Using machine learning to replicate chaotic attractors and calculate
  lyapunov exponents from data.
\newblock {\em Chaos: An Interdisciplinary Journal of Nonlinear Science},
  27(12), 2017.

\bibitem[Que24]{querlioz2024physics}
D.~Querlioz.
\newblock Physics solves a training problem for artificial neural networks.
\newblock {\em Nature}, 632(8024):264--265, 2024.

\bibitem[R{\"o}s76]{rossler1976equation}
O.E. R{\"o}ssler.
\newblock An equation for continuous chaos.
\newblock {\em Physics Letters A}, 57(5):397--398, 1976.

\bibitem[Sak88]{sakaguchi1988cooperative}
H.~Sakaguchi.
\newblock Cooperative phenomena in coupled oscillator systems under external
  fields.
\newblock {\em Progress of theoretical physics}, 79(1):39--46, 1988.

\bibitem[SK86]{sakaguchi1986soluble}
H.~Sakaguchi and Y.~Kuramoto.
\newblock A soluble active rotater model showing phase transitions via mutual
  entertainment.
\newblock {\em Progress of Theoretical Physics}, 76(3):576--581, 1986.

\bibitem[SLP22]{shougat2022dynamic}
M.R.E.U. Shougat, X.F. Li, and E.~Perkins.
\newblock Dynamic effects on reservoir computing with a hopf oscillator.
\newblock {\em Physical Review E}, 105(4):044212, 2022.

\bibitem[Str00]{strogatz2000kuramoto}
S.H. Strogatz.
\newblock From kuramoto to crawford: exploring the onset of synchronization in
  populations of coupled oscillators.
\newblock {\em Physica D: Nonlinear Phenomena}, 143(1-4):1--20, 2000.

\bibitem[SVM07]{sanders2007averaging}
J.A. Sanders, F.~Verhulst, and J.~Murdock.
\newblock {\em Averaging methods in nonlinear dynamical systems}, volume~59.
\newblock Springer, 2007.

\bibitem[TKK{\etalchar{+}}23]{tsunegi2023information}
S.~Tsunegi, T.~Kubota, A.~Kamimaki, J.~Grollier, V.~Cros, K.~Yakushiji,
  A.~Fukushima, S.~Yuasa, H.~Kubota, K.~Nakajima, et~al.
\newblock Information processing capacity of spintronic oscillator.
\newblock {\em Advanced Intelligent Systems}, 5(9):2300175, 2023.

\bibitem[TRAA{\etalchar{+}}17]{torrejon17}
J.~Torrejon, M.~Riou, F.~Abreu~Araujo, S.~Tsunegi, G.~Khalsa, D.~Querlioz,
  P.~Bortolotti, V.~Cros, K.~Yakushiji, A.~Fukushima, H.~Kubota, S.~Yuasa, M.D.
  Stiles, and J.~Grollier.
\newblock Neuromorphic computing with nanoscale spintronic oscillators.
\newblock {\em Nature}, 547:428, 2017.

\bibitem[TTN{\etalchar{+}}19]{tsunegi2019physical}
S.~Tsunegi, T.~Taniguchi, K.~Nakajima, S.~Miwa, K.~Yakushiji, A.~Fukushima,
  S.~Yuasa, and H.~Kubota.
\newblock Physical reservoir computing based on spin torque oscillator with
  forced synchronization.
\newblock {\em Applied Physics Letters}, 114(16), 2019.

\bibitem[VAL21]{verzelli2021learn}
P.~Verzelli, C.~Alippi, and L.~Livi.
\newblock Learn to synchronize, synchronize to learn.
\newblock {\em Chaos: An Interdisciplinary Journal of Nonlinear Science},
  31(8), 2021.

\bibitem[VLAA{\etalchar{+}}17]{vodenicarevic2017nanotechnology}
D.~Vodenicarevic, N.~Locatelli, F.~Abreu~Araujo, J.~Grollier, and D.~Querlioz.
\newblock A nanotechnology-ready computing scheme based on a weakly coupled
  oscillator network.
\newblock {\em Scientific reports}, 7(1):44772, 2017.

\bibitem[VLGQ18]{vodenicarevic2018nano}
D.~Vodenicarevic, N.~Locatelli, J.~Grollier, and D.~Querlioz.
\newblock Nano-oscillator-based classification with a machine
  learning-compatible architecture.
\newblock {\em Journal of Applied Physics}, 124(15), 2018.

\bibitem[Win67]{winfree1967biological}
A.T. Winfree.
\newblock Biological rhythms and the behavior of populations of coupled
  oscillators.
\newblock {\em Journal of theoretical biology}, 16(1):15--42, 1967.

\bibitem[Win01]{winfree1980geometry}
A.T. Winfree.
\newblock {\em The geometry of biological time}, volume~2.
\newblock Springer, 2001.

\bibitem[WS98]{watts1998collective}
D.J. Watts and S.H. Strogatz.
\newblock Collective dynamics of "small-world" networks.
\newblock {\em {N}ature}, 393(6684):440--442, 1998.

\bibitem[WYG{\etalchar{+}}19]{weng2019synchronization}
T.~Weng, H.~Yang, C.~Gu, J.~Zhang, and M.~Small.
\newblock Synchronization of chaotic systems and their machine-learning models.
\newblock {\em Physical Review E}, 99(4):042203, 2019.

\bibitem[ZGF{\etalchar{+}}23]{zuo2023self}
Z.~Zuo, Z.~Gan, Y.~Fan, V.~Bobrovs, X.~Pang, and O.~Ozolins.
\newblock Self-evolutionary reservoir computer based on kuramoto model.
\newblock {\em arXiv:2301.10654}, 2023.

\end{thebibliography}

\end{document}